\documentclass[10pt]{article} 
\usepackage{graphicx}
\usepackage[preprint]{tmlr}

\usepackage{amsmath,amsfonts,bm}

\def\eqref#1{equation~\ref{#1}}

\def\1{\bm{1}}

\def\vs{{\bm{s}}}

\DeclareMathAlphabet{\mathsfit}{\encodingdefault}{\sfdefault}{m}{sl}
\SetMathAlphabet{\mathsfit}{bold}{\encodingdefault}{\sfdefault}{bx}{n}

\DeclareMathOperator*{\argmax}{arg\,max}

\usepackage{wrapfig}
\usepackage{amsmath}
\usepackage{amsfonts}
\usepackage{amssymb}

\usepackage{cancel}
\usepackage{array}
\usepackage{multirow}
\usepackage{subcaption}
\usepackage{booktabs}
\usepackage{array}
\usepackage{threeparttable}
\usepackage{hyperref}
\usepackage{cleveref}
\usepackage{authblk}
\usepackage{graphbox}
\usepackage{enumitem}
\newcolumntype{L}[1]{>{\raggedright\arraybackslash}p{#1}}

\newcommand\HFWNumMachines{43}
\newcommand\NumMachinesInUse{19}

\newcommand\NumWeaponsInUse{6}

\newcommand\NumTrapsInLoadout{10}

\newcommand\NumElementalsInUse{7}
\newcommand\NumPartRemovalsInUse{3}

\newcommand\NumStylesInUse{20}

\newcommand\catrac{Cat-RAC} 
 
\newcommand\playstyles{playstyles}

\usepackage{mathtools}

\DeclarePairedDelimiter{\set}{\{}{\}}

\newcommand{\simplex}{\triangle}

\DeclareRobustCommand\onedot{\futurelet\@let@token\@onedot}
\def\@onedot{\ifx\@let@token.\else.\null\fi\xspace}
\def\eg/{\emph{e.g}\onedot} \def\Eg/{\emph{E.g}\onedot}
\def\ie/{\emph{i.e}\onedot} \def\Ie/{\emph{I.e}\onedot}
\def\cf/{\emph{c.f}\onedot} \def\Cf/{\emph{C.f}\onedot}
\def\vs/{\emph{vs}\onedot} \def\Vs/{\emph{Vs}\onedot}
\def\etc/{\emph{etc}\onedot}
\def\wrt/{with respect to} \def\dof/{d.o.f\onedot}
\def\etal/{\emph{et al}\onedot}
\def\viceversa/{\emph{vice-versa}}
\def\ow/{\emph{o.w}\onedot}
\def\whp/{w.h.p\onedot}
\def\apriori/{\emph{a priori}} \def\Apriori/{\emph{A priori}}
\def\ala/{\`{a} la}
\def\naive/{na\"{\i}ve} \def\Naive/{Na\"{\i}ve}
\def\GranTurismo7/{Gran Turismo\textsuperscript{\texttrademark} 7}
\def\PlayStation5/{PlayStation 5}
\def\GTSophy/{the GT agent}  

\def\gaussianMean{\mu}
\def\gaussianStd{\sigma}
\def\gaussian{\mathcal{N}}

\usepackage{hyperref}
\usepackage{url}

\title{Coachable Agents for Interactive Gameplay}

\author[2,$\dagger$]{Roberto Capobianco}
\author[1,$\ddagger$]{Harm van Seijen}
\author[1]{Nolan D. Bard}
\author[4]{Neil Burch}
\author[1]{Fatima Davelouis}
\author[1]{Josh Davidson}
\author[2]{Alisa Devlic}
\author[4]{Yunshu Du}
\author[1]{Ishan Durugkar}
\author[1]{Siddhant Gangapurwala}
\author[1]{Daniel Hernandez}
\author[1]{G. Zacharias Holland}
\author[1]{Sahil Jain}
\author[3]{Kenta Kawamoto}
\author[1]{Raksha Kumaraswamy}
\author[1]{Patrick MacAlpine}
\author[4]{Dustin R. Morrill}
\author[1]{Declan Oller}
\author[2]{Francesco Riccio}
\author[4]{Akanksha Saran}
\author[4]{Craig Sherstan}
\author[2]{Kaushik Subramanian}
\author[1]{Thomas J. Walsh}

\author[1]{Samuel Barrett}
\author[1]{Kizza N. Frisbee}
\author[1]{Mady Govil}
\author[1]{Johannes G\"{u}nther}
\author[1]{Varun R. Kompella}
\author[1]{James A. MacGlashan}
\author[1]{Maxwell Svetlik}
\author[1]{Michael D. Thomure}
\author[1]{Jaden B. Travnik}
\author[1]{Kevin Waugh}

\author[4]{Elahe Aghapour}
\author[4]{Florian Fuchs}
\author[4]{Andreanne Lemay}
\author[4]{Shruti Mishra}
\author[4]{Takuma Seno}

\author[1]{Peter Stone}
\author[4]{Michael Spranger}
\author[4]{Peter R. Wurman}

\affil[1]{Sony Corporate Technology Center America, Inc.}
\affil[2]{Sony Europe Ltd.}
\affil[3]{Sony Group Corporation}
\affil[4]{Work done while at Sony}
\affil[$\dagger$]{roberto.capobianco@sony.com}
\affil[$\ddagger$]{harm.van.seijen@sony.com}

\def\month{10}  
\def\year{2026} 
\def\openreview{\url{https://openreview.net/forum?id=GMPy1cDgml}} 

\begin{document}

\maketitle

\begin{abstract}
Reinforcement learning has proven to be a valuable tool in the creation of advanced AI and robotic systems, contributing to everything from game playing~\citep{Silver2016AlphaGo, Silver2017AlphaZero,wurman2022outracing} to robotics~\citep{levine2016end,andrychowicz2020learning,luo2025precise, durr2026outplaying} to foundation models~\citep{ouyang2022training,noukhovitchasynchronous,sun2024aligning,guo2025deepseek}.
Through trial and error, these AI systems typically learn a single, near-optimal behavior to solve their tasks.
However, there are many use cases in which one would like to assert some level of control, after training is completed, over \emph{how} the task is solved. 
We refer to these modifications of a core task as \emph{styles}.
We combine universal value function approximators (UVFAs) with carefully selected training scenarios, learning algorithms, and data augmentation to create a framework for coaching agents that exhibit styles in complex domains. 
We demonstrate the framework's application in the AAA video games Horizon Forbidden West and Gran Turismo, and in an open-source humanoid test domain.
Despite the different nature of the domains---car racing, stylized game combat, and humanoid walking---each agent shows strong coherence to the style requests while still satisfying the main task in its domain.
Importantly, the techniques outlined in this paper allow an end user to choose the final behavior at run time, giving them flexible control over the final executed performance.
\end{abstract}

\section{Introduction}

With artificial intelligence (AI) entering the mainstream, there is a growing focus on enabling users to meaningfully align the system's behavior to the user's preferences. 
While users are now familiar with requesting generative AI tools~\citep{riemer2024conceptualizing,asperti2025critical} to follow style requests (``generate an impressionist painting of a cat playing table tennis.''), such control has so far been unavailable for agents operating in real-time control settings. 
Concurrently, reinforcement learning (RL) has become a leading method for achieving state-of-the-art performance across a wide range of real-time control domains~\citep{wurman2022outracing,kaufmann2023champion,hwangbo2017control}.
These RL agents typically master a single task and perform it in a single, near-optimal way. 
In practice, in addition to \textit{what} an agent accomplishes, users often care about \textit{how} it accomplishes the task, which may depend on the context at runtime.
For example, we might prefer our cleaning robot to clean up the room \textit{quietly} because the baby is sleeping, or to prioritize \textit{speed} because guests are about to arrive. 
These preferences do not change the underlying task itself---cleaning the room---but they do express valuable, context-dependent preferences over the agent's behavior.

\begin{figure}[t]
\centering
\includegraphics[width=0.9\linewidth]{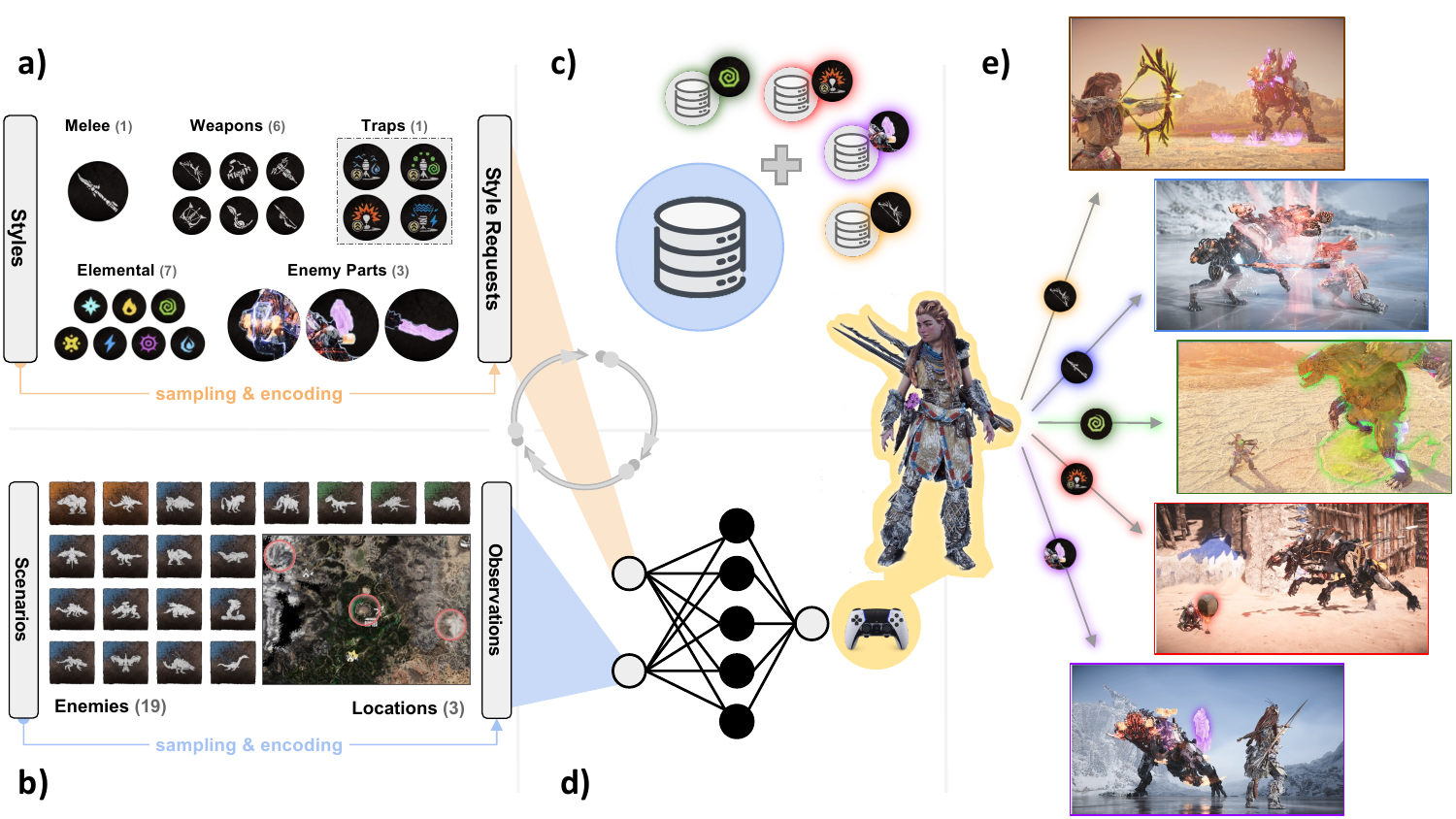}
\caption{
\textbf{Coachable agents overview:} coaching an agent that could play parts of the game Horizon Forbidden West involved creating training scenarios that rewarded different \textbf{a} combat styles against \textbf{b} \NumMachinesInUse\ different enemy machines in three different locations in the game world. The trajectories, stored in \textbf{c} replay buffers (including tables for specific styles), are used to train \textbf{d} a policy that takes in observations of the world and style request signals and outputs game controller actions. By iterating through this loop, the policy improves and generates better and better training data. The resulting policy is able to \textbf{e} fight any of the machines in any of the three locations while exhibiting a high level of adherence to the style requests.
}
\label{fig:panel-1}
\end{figure}

Coachable agents are trained to perform a task in a variety of different \textit{styles}, and to apply those styles on-demand at execution time. 
This problem framing differs from multi-task~\citep{kalashnikov2021mt,sutton2011horde} or goal-conditioned learning~\citep{liu2022goal} in which the problem statement is focused on teaching an agent to accomplish multiple different tasks.  To understand the difference, consider the robot cleaning scenario: classifying ``cleaning quietly'' and ``cleaning quickly'' as two different tasks downplays their commonality and obscures the fact that there is a spectrum of ways to be quiet and/or quick that the user may want to select from.
With styles---similar to how an adverb modifies a verb---we keep the primary task constant but control the secondary performance characteristics.

Much like coaching human athletes, effective agent coaching centers on designing the incentives and practice scenarios required to acquire domain-specific skills and style variations. In this study, we explore these concepts across three domains, each presenting a unique set of challenges.
The simplest environment is the humanoid model from the DeepMind Control Suite (DMC)~\citep{tassa2018deepmindcontrolsuite}, a standard RL test environment.\footnote{Source code of our framework applied to the humanoid domain is here: \href{https://github.com/sonyresearch/coachable_agents}{https://github.com/sonyresearch/coachable\_agents}.} 
The other two domains are AAA video games: the racing game \GranTurismo7/ (GT) and the open-world adventure game Horizon Forbidden West (HFW).
Modern commercial video games present richly interactive worlds with well-defined rules and clear success metrics, making them excellent proving grounds for advances in AI~\citep{Vinyals:2019,kanervisto2025world}.
Moreover, video games are often designed to allow players to play with a variety of different \playstyles~\citep{bartle1996hearts}, making them ideal test environments.

In this work, we describe the innovations necessary to scale to the complexity of modern video games, and suggest best practices for training coachable agents. Figure~\ref{fig:panel-1} illustrates the main components of the coaching framework in the context of HFW.


\section{Style-Conditioned UVFA}
\label{sec:style-conditioned UVFA}

The foundation of our framework for coaching agents is reinforcement learning, the framework in which software agents learn a task through trial and error~\citep{Sutton1998}.
The field gained broad public attention around 2016 when RL, combined with deep neural networks, was fundamental to several high-profile AI milestones \citep{mnih2013Atari,Silver2016AlphaGo}.
RL algorithms usually make use of a \textit{policy}, $\pi(s): s \rightarrow a$, and/or a \textit{critic}, $Q_{\pi}(s,a): s, a \rightarrow V_{\pi}$, where $V_{\pi}$ approximates the expected cumulative (usually discounted) future reward of taking action $a$ in state $s$ and following $\pi$ thereafter.
To make RL work in practice, designers must choose among a variety of algorithms, environment state representations, reward formulations, and training paradigms to produce an agent that can achieve the desired outcome~\mbox{\citep{ghasemi2025comprehensivesurveyRL}}. 
The standard application of RL often leads to agents that perform their tasks well but offer little runtime flexibility as to how those tasks are carried out. 

To train coachable agents we build on prior work on universal value function approximators (UVFAs)~\citep{HORNIK1989359,pmlr-v37-schaul15}, which extend the standard value function to condition on ``goal states''. 
Unlike the original UVFA formulation, our approach addresses scenarios in which the core objective remains fixed, but multiple valid behaviors exist for accomplishing it. 
We refer to this variation as \textit{style-conditioned UVFA}. In this version of UVFA, the reward function consists of two parts: a part that captures the task objective and remains fixed throughout training, and a part that pushes the agent towards certain behavioral styles and is conditioned on a style-parameter vector $\boldsymbol{\theta}$ that can be varied on the fly at runtime:
$$r(s,a; \boldsymbol{\theta}) = r_{task}(s,a)  + r_{styles}(s,a;\boldsymbol{\theta})\,.$$
For the main experiments in this paper, we use a linear form for $r_{styles}$:
$$r_{styles}(s,a;\boldsymbol{\theta}) = \sum_{i=1}^N \theta_i\, r_{style\_i}(s,a)$$
with $\boldsymbol{\theta} = (\theta_1, \theta_2, \dots, \theta_N)^\top$ encoding the linear weights of $N$ separate style-reward functions, $r_{style\_i}$, each representing one distinct style. Just like the goal vector in standard UVFA, the style parameter is appended to the observation, becoming part of the input for the actor and critic networks. 
Besides acting as linear weights to specific style-reward functions, style parameters can influence a reward function in other ways, such as acting as thresholds (for gating/ramping) or targets (setpoints) that penalize deviation from
desired ranges. For example, in GT, we use two of the style-parameter values as thresholds to set a variable-sized penalty region around opponents. This can be used to modulate \GTSophy/'s apparent aggressiveness (see Figure \ref{fig:lat-control} in Appendix \ref{app: extended data}). Similarly, in DMC one style-parameter value is used to set the gait (i.e., the step length) that the humanoid should aim for.


Style-conditioned UVFA is related to multi-objective learning~\citep{abels2019dynamic,hayes2021practical}, which aims to optimize a vector of multiple, often conflicting reward functions simultaneously. A key distinction is that style-conditioned UVFA maintains a permanent emphasis on one overarching task objective.
Another distinction is that while many distinct styles can be learned using a single model (in HFW, we learn up to 20), typically only one or two are active at the same time. That said, similar to multi-objective learning, following a certain style \emph{can} reduce task performance. For example, in GT, being careful about tire-wear will reduce the pace, or in HFW, fighting a strong enemy with a weak weapon will decrease the win-rate. How to balance task performance with adherence to a style is ultimately a question for the user to answer. To support this, we train our agents across a range of ${\boldsymbol{\theta}}$-values that yield different trade-offs between task and style performance. This enables a user to tune the trade-off to their liking after training. Specifically, in the linear formulation above, each $\theta_i$ has a range $[\theta_{i,min},\, \theta_{i,max}]$ and we train across values in this range. By convention, a positive $\theta_i$ encourages a style and $\theta_{i,max} > 0$ for all styles. In most cases, we use $\theta_{i,min} = 0$, but in some cases extending into the negative range can yield new interesting behaviors. In GT, the tire-wear style applies a penalty for tire-wear, encouraging the agent to be careful with its tires. Using a negative weight for this style leads to a drifting behavior, a driving technique where the car intentionally slides through corners by breaking rear-wheel traction (see Appendix \ref{app: videos}  for a video). And in HFW, we can use negative weights to discourage the use of a particular weapon (see the ``No Skykiller'' style in Figure \ref{fig:HFW_style_vs_damage}).

The idea of conditioning on diverse reward functions is not entirely new. For instance, \cite{de2021configurable, de2022automated} demonstrated that carefully engineered reward terms can incentivize agents to learn multiple policies, enabling distinct playstyles in a small, turn-based video game. And more recently, \cite{nauman2026rewardconditioned} applied reward-conditioning via off-policy learning on data generated from a primary task reward. However, their resulting policy generalizes poorly to these supplementary rewards without fine-tuning, and their evaluation was restricted to small domains like OpenAI Gym and DMC Vision.  Our work elevates reward-conditioning to high-dimensional control domains, a substantially more challenging setting. 


\section{Method}

Our method relies on an off-policy actor-critic backbone with replay. We use
numerous asynchronous rollout workers that pull the latest policy networks and sample a style-parameter vector at episode start (which remains fixed within an episode), collect data and stream transitions to a shared replay buffer.  A single learning process updates the actor/critic. Episodes end on success/failure (e.g., all opponents defeated) or on a fixed time limit. Style conditioning follows the UVFA formulation described in Section \ref{sec:style-conditioned UVFA}. As in standard RL, we must choose the state features, rewards, and action representations for the agent to have the capacity to succeed in each domain. 
Because our experimental domains differ widely in complexity, dynamics, and action spaces, we use different algorithmic variations per environment, summarized in Table~\ref{tab:domain_training_summary} of Appendix \ref{app: extended data}.


\subsection{Base Algorithm}

As the base algorithm, we use different variations of Soft Actor Critic (SAC). For the humanoid domain, we follow the canonical SAC formulation and perform \emph{simultaneous} actor/critic updates each step using batches from the replay buffer, while the entropy temperature is fixed.
For \GranTurismo7/ we adopt the quantile-regression variant of SAC (QR-SAC) from \cite{wurman2022outracing}, replacing scalar Q with a distributional critic trained by quantile regression while keeping the SAC structure (stochastic policy, replay, target networks). 

Horizon Forbidden West control mixes discrete button combinations with continuous sticks/triggers. To support mixed action spaces of this kind we generalize SAC to a \emph{regularized actor–critic} (RAC) objective that replaces the entropy term with a generic regularizer $\mathcal{R}(a\mid s,f(\boldsymbol{\theta}))$. This modification lets us apply different regularizers per action subspace and sum them in a single objective while retaining the SAC training loop. Specifically, we apply \emph{exact} categorical entropy to the discrete policy heads and an \emph{n}-sample empirical mean of the negative log density from the $\tanh$-Gaussian continuous policy heads. In our experiments, we estimate the negative entropy by averaging $n=10^3$ samples per component, instead of the conventional single sample. This approach is substantially more accurate than the single-sample surrogate commonly used in SAC, without a meaningful slowdown in practice. Using exact categorical entropy is necessary because gradients cannot be propagated through categorical samples. Further details of our RAC implementation can be found in Appendix \ref{app: training algorithms}.

\subsection{Task Impersonation}
Task impersonation~\citep{kalashnikov2021task-impersonation} is a relabeling technique that improves data efficiency by reusing the same transitions under alternative ``task'' descriptors. We adopt this approach in HFW, and in our setting there are no discrete tasks; instead, each style-parameter vector $\boldsymbol{\theta}$ specifies a distinct style-reward configuration. We therefore treat sampled style settings as impersonated tasks: a single transition can be evaluated under multiple $\boldsymbol{\theta}$'s without altering the underlying dynamics.

As discussed above, we sample style parameters from a certain distribution. During updates, however, we increase coverage over $\Theta$ by drawing additional parameter vectors i.i.d.\ from the same distribution (i.e., $\hat{\boldsymbol{\theta}}^{(j)}\!\sim\!p_{\mathrm{\boldsymbol{\theta}}}$).
In effect, task impersonation provides a more accurate Monte Carlo approximation of our objective's expectation over $\boldsymbol{\theta}$, but without modifying the replay buffer: every transition sampled from the replay buffer is paired, on each step, with several freshly sampled UVFA configurations, expanding the style space coverage.

Specifically, each update proceeds by sampling a mini-batch of transitions $(s_t,a_t,s_{t+1})$ from the replay buffer, retaining the label defined by the rollout parameters $\boldsymbol{\theta}^{(0)}$, and generating $n$ auxiliary labels by recomputing the style rewards under $\hat{\boldsymbol{\theta}}^{(1)},\ldots,\hat{\boldsymbol{\theta}}^{(n)}$.
When forming the new update targets, we condition the networks on the corresponding encodings $f(\boldsymbol{\theta})$.
New relabeled data is created at the start of each training step and discarded at the end, and is never written to the replay buffer.
Because each update processes $(n{+}1)$ labeled versions of every transition, throughput decreases with $n$; empirically we found $n=4$ to be a good trade-off in wall-clock time (see Section \ref{sec:ablations} for some ablations).

A design choice worth noting is that impersonated labels are neither filtered nor re-balanced: styles act through rewards and conditioning features rather than the environment dynamics, so most transitions remain informative---positively or negatively---for most style configurations. Averaging contributions uniformly across the $(n{+}1)$ labels aligned well with the objective function’s expectation over $\boldsymbol{\theta}$ and worked reliably in our experiments.

\subsection{Stratified Sampling}
In GT and HFW, we additionally use stratified sampling via \emph{multi-tables}~\citep{wurman2022outracing} and its successor, \emph{event tables}~\citep{kompella2023eventtables}, respectively. Multi-tables partition the replay buffer into separate tables that capture different portions of trajectories, grouped by the features and scenarios they represent, enabling balanced learning across the situations the agent encounters. Event tables organize replay data by style configuration and promote learning across style settings. For each style dimension $i \in {1,\dots,N}$, 
trajectories generated under a style-parameter vector $\boldsymbol{\theta}$ are inserted into the event table associated with that style configuration. Transitions are stored only when a style-specific event is triggered (e.g., damage attributable to that style or part removal), with a maximum event horizon $H_{\max}$ representing the number of steps stored per event. 

For both multi- and event tables, each table has a fixed capacity and is updated in a first-in, first-out manner. During training, mini-batches are sampled in a stratified manner across these tables, ensuring balanced coverage of scenarios or contrasting style settings, and preventing over-representation of dominant regimes.
Unlike task impersonation, which relabels transitions with alternative style parameters to expand style coverage, these approaches operate at the data organization level, preserving the original labels while promoting balanced sampling across scenarios and styles.

\subsection{Style-Parameter Sampling}

For GT, we implement five different style components to control overtaking, defending, tire saving, fuel saving, and aggressiveness. The style parameters for the first 4 styles encode linear weights; the aggressiveness style uses 2 non-linear values. All style parameters are drawn independently from different distributions. For full details, see Appendix \ref{app: domain details}.

For HFW, we use 18 different style components with 18 linear weights. Due to the high number of styles and many of them being incompatible with each other, independent sampling as done in GT results in poor performance (see ablations in Section \ref{sec:ablations}). Hence, we mainly follow a ``continuous one-hot'' sampling strategy. This means that $\theta_i = 0$ for all styles, except for one, the active style, whose value is drawn uniformly from its range. The active style itself is also drawn uniformly at random from the 18 styles.

For the Humanoid domain, we implemented 2 different styles, one related to arm poses and one related to the step length. The style parameter for the arm poses encodes which style pose should be used, while the one for the step length directly encodes the length of the step. The specific values for the 2 styles are drawn independently from each other.

\section{The Test Domains}


GT represents a realistic, real-time control scenario in which the slightest error can lead to catastrophe. The styles we present for GT reflect real-world driving behaviors, such as the strategic decision to sacrifice lap time to delay a pit stop.

Because the motifs of the game HFW drive our style choices, we aim to provide the reader with enough background to understand the results.
In this open-world game, the heroine, Aloy, battles a variety of animal-inspired robots across a post-apocalyptic version of the American West.
Aloy always carries a spear for melee combat, and, through a \textit{weapon wheel}, the player has quick access to six projectile weapons, each having two or three ammo types.
Aloy can also lay proximity traps and lure machines into them, and she can consume healing berries to recover health.
All of this flexibility requires extensive use of the 14 boolean buttons, two 2D joysticks, and the two continuous triggers on the PlayStation 5 controller.
The enemy machines are complex non-player characters (NPCs) with dozens of animated attack types, many of them unique to each machine type. 
They typically have components---including weapons, shields, and canisters---that can be targeted and removed.
Aloy's ammo and traps often carry an \textit{elemental type} that can build up and put the target into an \textit{elemental state}.
For example, if Aloy hits a machine with several fire arrows, the machine may reach a burning state in which it takes continuous fire damage for a period of time.
Table~\ref{tab:machineDetails} in Appendix \ref{app: extended data} shows details of the machines in the training set, and we refer interested readers to any number of online resources with extensive information on the game.
The styles our agent exhibits in HFW reflect the breadth of combat choices the developers created for players, allowing them to choose their favorite weapons, play it safe with traps, or leverage the powerful elemental systems of the game.

\begin{figure}[t]
    \centering
    \includegraphics[width=1.00\textwidth]{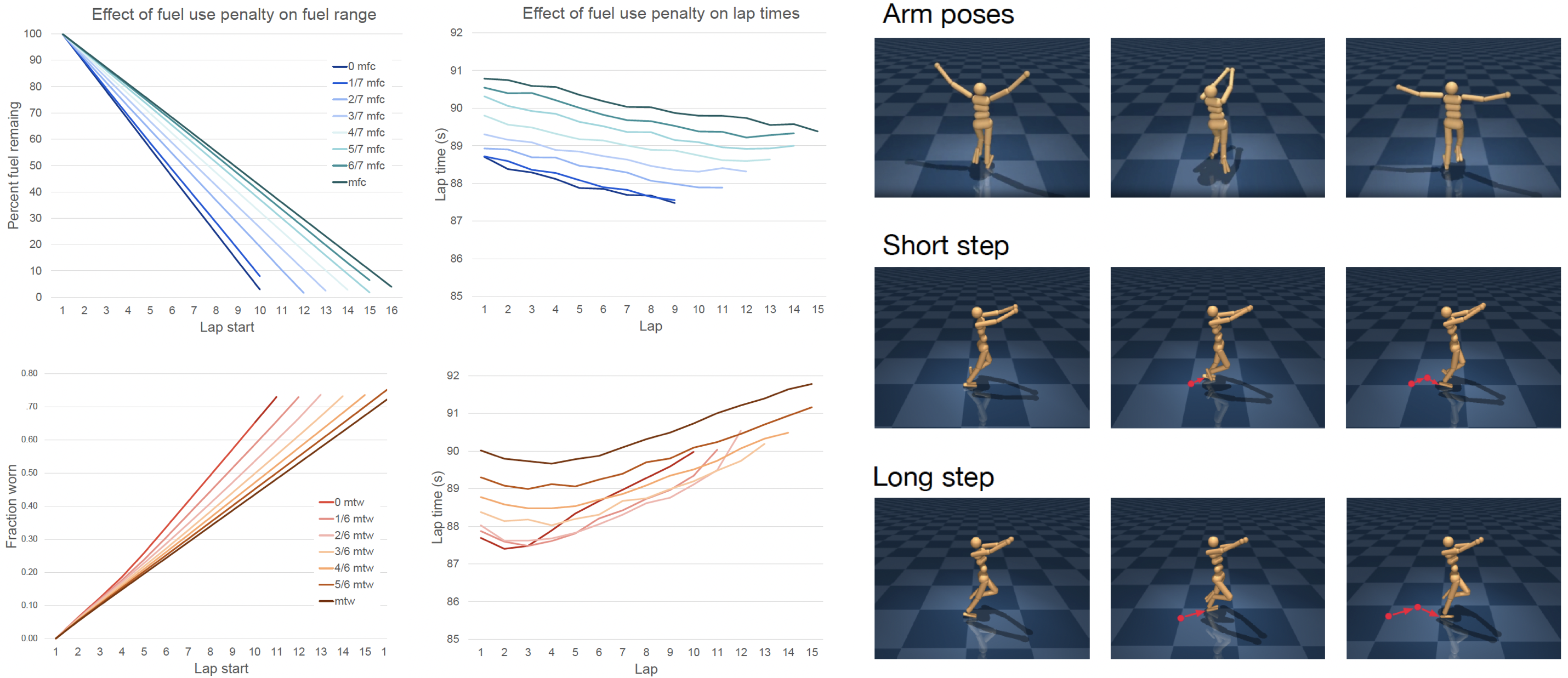}
    \caption{(left) \textbf{GT Style Performance:}
Style-conditioned UVFA was used to create a version of \GTSophy/ that can trade off lap time for fuel range and/or tire life. At the environment level, Gran Turismo 7 allows us to turn fuel consumption or tire wear on or off independently, allowing us to study each situation in isolation.  In the plots, \emph{mfc} is the max fuel consumption penalty and \emph{mtw} is the max tire wear penalty used in training each variant of GT Sophy. The data was collected using the Gran Turismo Red Bull X2019 Competition race car on Lago Maggiore - Full course (GP) with accelerated tire wear or fuel consumption.  (right) \textbf{Humanoid Style Examples:} Three arm poses of the humanoid domain and the extremes of the parameterized gait length control. The agent is able to combine any of the three arm poses with any gait
length between 0.3m and 0.65m.
}
\label{fig:GT_and_humanoid_domain}
\end{figure}

To establish the generality of our approach beyond games, we also train an agent on the DeepMind Control Suite~\citep{tassa2018deepmindcontrolsuite} humanoid model. 
The humanoid is an unstable dynamic system with a 21-dimensional action space whose core objective is to walk across an infinite plane.
On top of the base task we layer styles to control discrete arm poses that can be combined with continuous gait lengths, as shown in Figure~\ref{fig:GT_and_humanoid_domain}.
From this third domain, one can see how the approach could extend into many robotics applications.


\section{Results}

Our main empirical evaluation centers on the game Horizon Forbidden West. The large variety in game-play that this game enables makes it an ideal testbed for experimenting with a large number of different styles. But we first show results from the game Gran Turismo, which represents a very different type of game. For the humanoid domain, a video accessible through Appendix \ref{app: videos} provides qualitative results demonstrating strong adherence to the requested style.

\subsection{Gran Turismo}

We show the use of style-conditioned UVFA to influence aspects of \GTSophy/'s driving style in a pair of experiments to control tire wear and fuel usage (Figure~\ref{fig:GT_and_humanoid_domain}).
By adjusting the style parameters, the user can trade off pace against fuel range or tire life, enabling control over the key strategic decision about when to take a pit stop. 
Plots \textbf{a} and \textbf{c} show how each penalty can be used to extend the time until a pit stop is needed by as much as 60\%. Graphs \textbf{b} and \textbf{d} show how maximally conserving these resources cost the agent 2-3 seconds per lap.

\subsection{Horizon Forbidden West}

For experiments in HFW, we configure the agent as a late-stage player in order to give us access to the widest variety of skills, weapons, and challenging opponents.
Thus, Aloy has maximum health (700 HP), the fully unlocked skill tree, upgraded armor, and six fully upgraded weapons.
Table~\ref{tab:weaponsDetails} in Appendix \ref{app: extended data} shows Aloy's weapons, and more details on Aloy's configuration and the agent's action space are available in Appendix \ref{app: domain details}. We provide an overview video for HFW, as well as videos of all the individual styles, that is accessible through Appendix \ref{app: videos}.

\begin{figure}[t]
    \centering
    \includegraphics[width=1.0\textwidth]{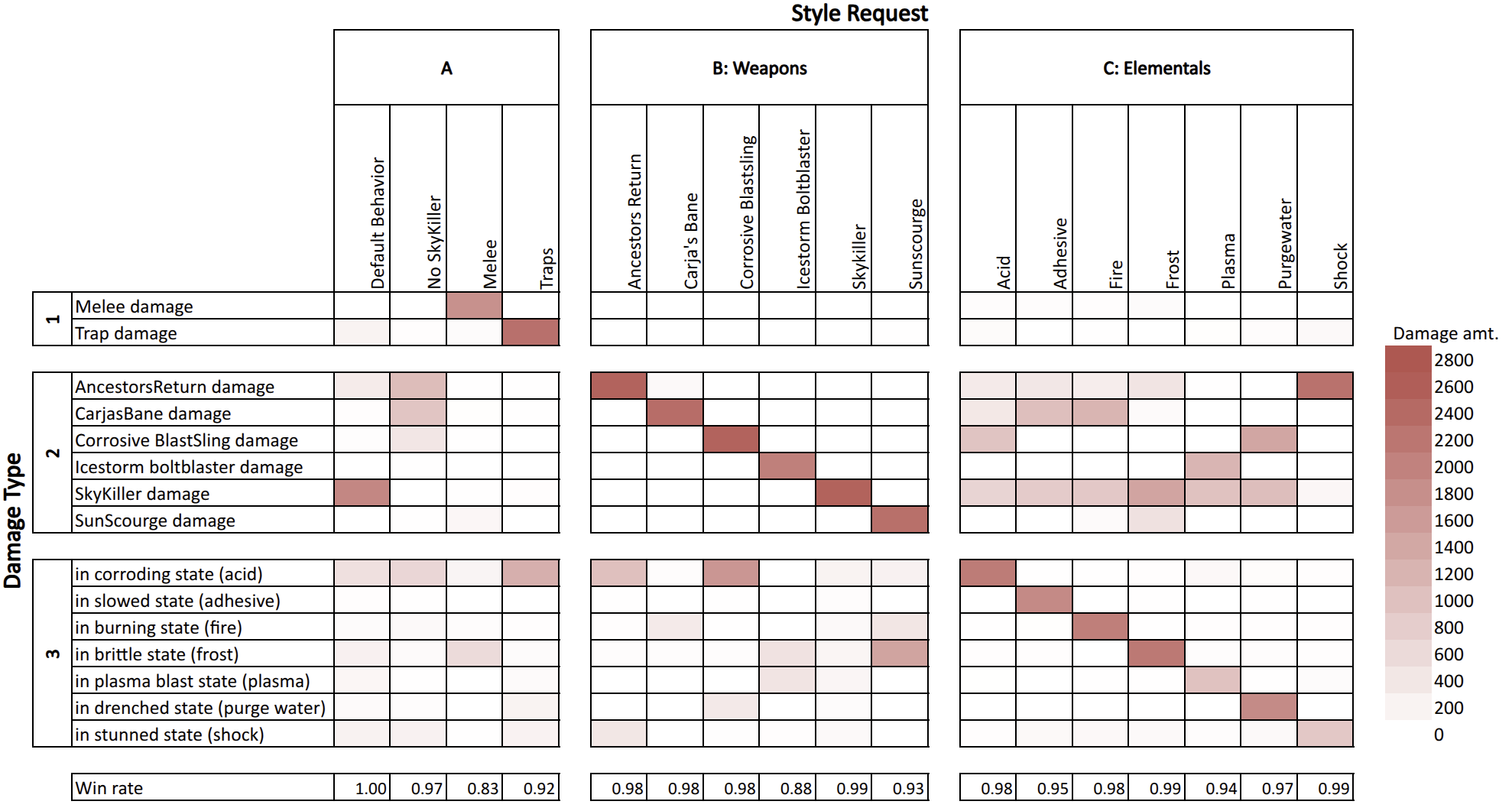}
    \caption{\textbf{HFW Style Performance:} The column groups A to C represent distinct styles that can be requested, and the rows are
damage metrics collected from fights against all 19 enemies. The task performance (i.e., the average win rate across enemies) is summarized in the last row. The intensity of the color in each cell represents the average
damage done across enemies. 
See Figure~\ref{fig:HFW-damage-done-by-style} in Appendix \ref{app: extended data} for a numerical version of this matrix.
The strong diagonal from melee to the shock elemental shows the main result: the high correlation between the style requested and the type of damage the agent inflicts.}
    \label{fig:HFW_style_vs_damage}
\end{figure}

\subsubsection{Style Performance}

Aligning with the game’s combat themes, we implemented 18 styles that encourage melee combat, traps, the 6 weapons configured in the weapon wheel, \NumElementalsInUse\ elemental styles, and \NumPartRemovalsInUse\ part-removal styles.  We also defined 2 additional styles. The first is the \textit{default style}, which disables style rewards ($\boldsymbol{\theta} = \boldsymbol{0}$) and reduces the objective strictly to the task reward. The second is the ``No Skykiller'' style, designed to illustrate that our framework can actively \textit{discourage} specific strategies. By testing both positive and negative reward weights on a dominant weapon (the Skykiller) we show how behavior can be restricted as easily as encouraged. This brings our total to 20 distinct styles.

We ran five randomly seeded trainings to a fixed 4.5 million gradient steps and evaluated the resulting five policies across \NumStylesInUse\ styles against \NumMachinesInUse\ enemies in three different map locations (see Figure~\ref{fig:panel-1}).
For each combination of machine-style-location-seed, we collected ten samples, resulting in data from 57,000 total battles. 
For all styles except the ``No Skykiller'' style, we set the corresponding UVFA weight at about 95\% of its max range-value when the style is activated; for the ``No Skykiller'' style, we set it at about 95\% of its min range-value (full range values can be found in Table \ref{tab:dynamic_rewards_hfw} of Appendix \ref{app: extended data}). 
For each battle, we recorded whether the agent or the machine won (win rate), what types of damage the agent inflicted on the machine (damage type), and the fraction of damage done that counts towards the requested style (style score). 

Figure~\ref{fig:HFW_style_vs_damage} shows the types of damage inflicted when different styles (in columns) are requested. 
The key result is that the diagonal elements dominate, showing a strong correspondence between the requested style and the damage type it is rewarded for. 
We also see that the agent's default style (column 1) favors the Skykiller weapon, but the agent also respects the request discouraging its use in the ``No Skykiller'' (column 2).

\begin{figure}[t]
    \centering
    \includegraphics[width=0.40\textwidth, align=c]{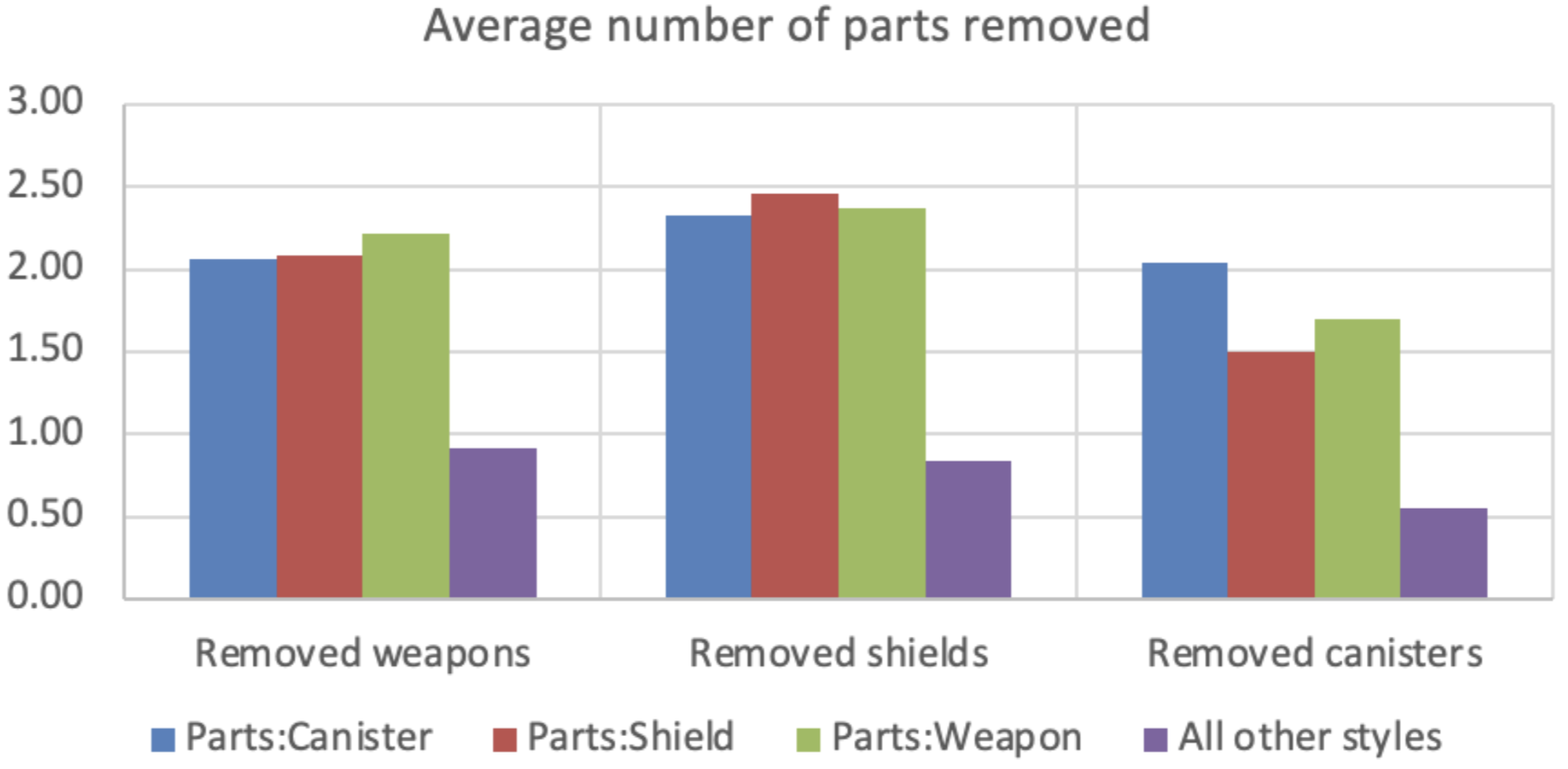}
    \includegraphics[width=0.58\textwidth, align=c]{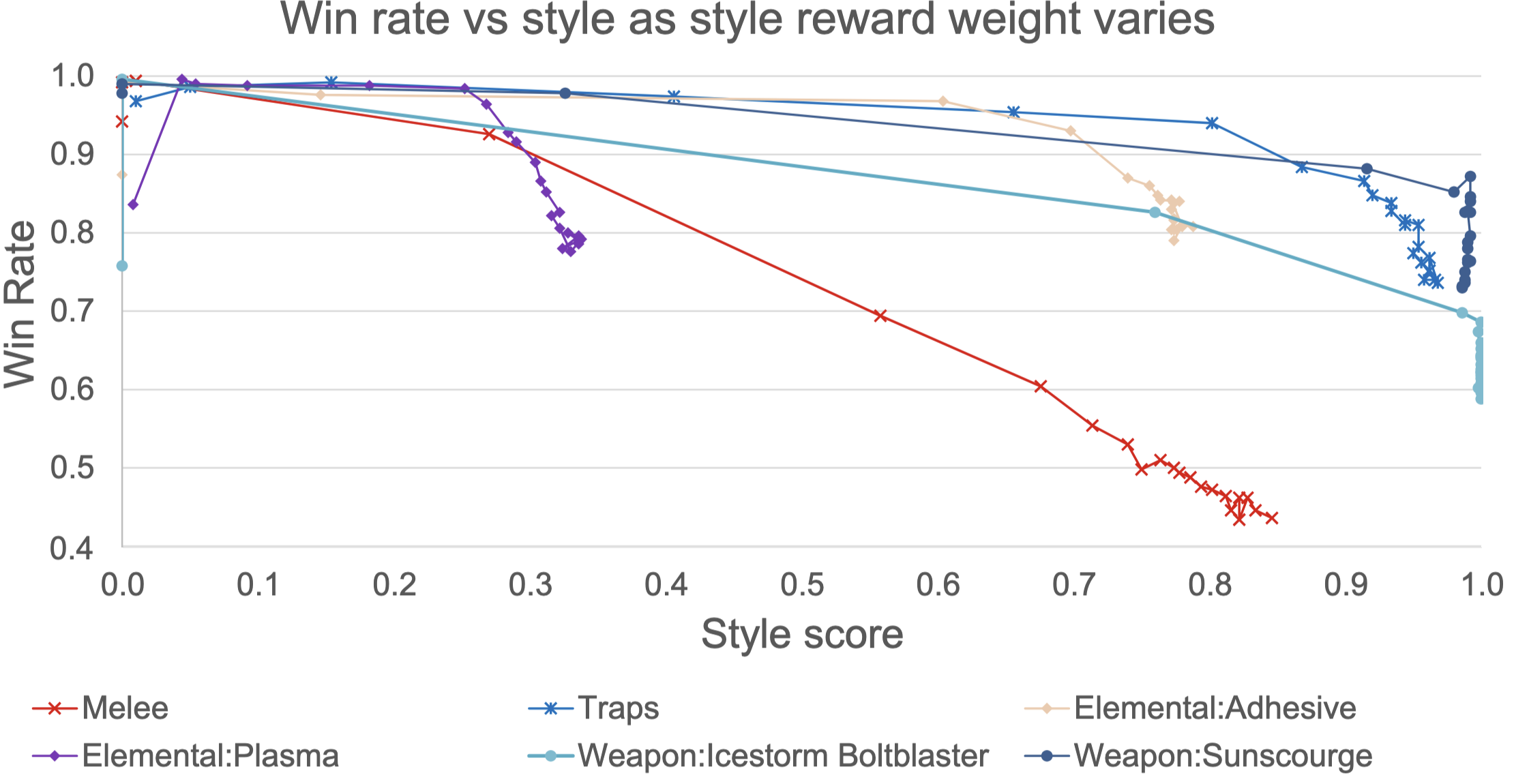}
    \caption{\textbf{[left]}
    Number of parts removed in response to the three part-removal styles. The agent removes more than 2x the number of parts when pursuing part removal strategies compared to other styles. The plots also show a strong correlation between the individual part-removal styles.
    \textbf{[right]}
    The win rate as a function of the style score as we scan
across the weight-range for six of the styles.
    }
    \label{fig:parts_removed_and_pareto}
\end{figure}

The 3 part-removal styles are scored by the number of machine components removed, rather than damage inflicted, so we separate those results into Figure~\ref{fig:parts_removed_and_pareto} (left).
When requested to, the agent removes about twice as many parts as all other styles.
However, because the agent was rewarded for ``removing'' the parts, regardless of whether they were detached or destroyed, the agent learned to use the Skykiller, with its explosive ammo, a crude strategy that often removes multiple parts at the same time.

Because some styles degrade task performance, users must balance task performance with style adherence. A major benefit of our style-conditioning approach is that users can fine-tune this trade-off post-training.
We quantify style adherence using a \emph{style score} that is bounded between 0 and 1. For part-removal styles, it is defined as the fraction of target enemy parts successfully removed relative to the enemy's total count for that part type. For all other styles, it is defined as the proportion of total damage dealt via the style's damage type. Figure \ref{fig:parts_removed_and_pareto} (right) shows the win rate versus style score for 6 distinct styles, generated by sampling the weight-range for each style at equidistant intervals. The result is a Pareto-like graph that can be used to decide what weight value to use for a particular style. For melee, the win rate can drop from close to 1.0 to 0.4, depending on how strongly the style request should be followed (in other words, how high its style weight is).

While Figures \ref{fig:HFW_style_vs_damage} and \ref{fig:parts_removed_and_pareto} show that overall the performance of the reward-conditioned UVFA framework is quite good, some of the styles have either a smaller win rate or show less adherence to the requested style (in Table \ref{tab:baseline style score win rate} of Appendix \ref{app: extended data} we show the style score and win rate of all styles).
An interesting question is how much better the agent would do if it was trained exclusively on that style. To investigate this, we selected five styles (three with a win-rate on the lower end; two with a style score that is small) and trained specialist agents for each of these. The results in Table \ref{tab:specialist agents} show that a specialist agent does not always outperform the UVFA agent. This can happen because the additional styles that are being learned with the UVFA framework can be viewed as auxiliary tasks to the requested style that yield better representations and/or increased exploration that ultimately outweighs any positive effects from dedicating all training cycles to one style.

\begin{table}[ht]
\centering
\caption{Performance of the reward-conditioned UVFA framework compared to a specialist agent that is trained exclusively on one style, for five different styles. Results are averaged over all 19 enemies, 3 locations, and 5 independent training runs.}
\label{tab:specialist agents}
\begin{tabular}{rl|cc|l|cc|}
\cline{3-4} \cline{6-7}
\multicolumn{1}{l}{}                                &  & \multicolumn{2}{c|}{Style score}          &  & \multicolumn{2}{c|}{Win rate}          \\ \cline{3-4} \cline{6-7} 
\multicolumn{1}{l}{}                                &  & \multicolumn{1}{c|}{UVFA}          & specialist        &  & \multicolumn{1}{c|}{UVFA}          & specialist        \\ \cline{1-1} \cline{3-4} \cline{6-7} 
\multicolumn{1}{|r|}{Melee}                         &  & \multicolumn{1}{c|}{0.90}          & \textbf{0.92} &  & \multicolumn{1}{c|}{0.83} & {\textbf{0.86}}          \\ \cline{1-1} \cline{3-4} \cline{6-7} 
\multicolumn{1}{|r|}{Traps}                         &  & \multicolumn{1}{c|} {0.98} & {\textbf{0.99}}          &  & \multicolumn{1}{c|}{0.92}          & \textbf{0.95} \\ \cline{1-1} \cline{3-4} \cline{6-7} 
\multicolumn{1}{|r|}{Weapons: Icestorm Boltblaster} &  & \multicolumn{1}{c|}{1.00} & 1.00          &  & \multicolumn{1}{c|}{0.88}          & \textbf{0.91} \\ \cline{1-1} \cline{3-4} \cline{6-7} 
\multicolumn{1}{|r|}{Elementals: Shock}             &  & \multicolumn{1}{c|}{0.41}          & \textbf{0.56} &  & \multicolumn{1}{c|}{\textbf{0.99}} & 0.71          \\ \cline{1-1} \cline{3-4} \cline{6-7} 
\multicolumn{1}{|r|}{Parts: Canister}               &  & \multicolumn{1}{c|}{\textbf{0.37}} & 0.34          &  & \multicolumn{1}{c|}{\textbf{0.96}} & 0.65         \\ \cline{1-1} \cline{3-4} \cline{6-7} 
\end{tabular}
\end{table}

\subsubsection{Runtime Style Switching}

\begin{figure}[t]
\centering
\includegraphics[width=\linewidth]{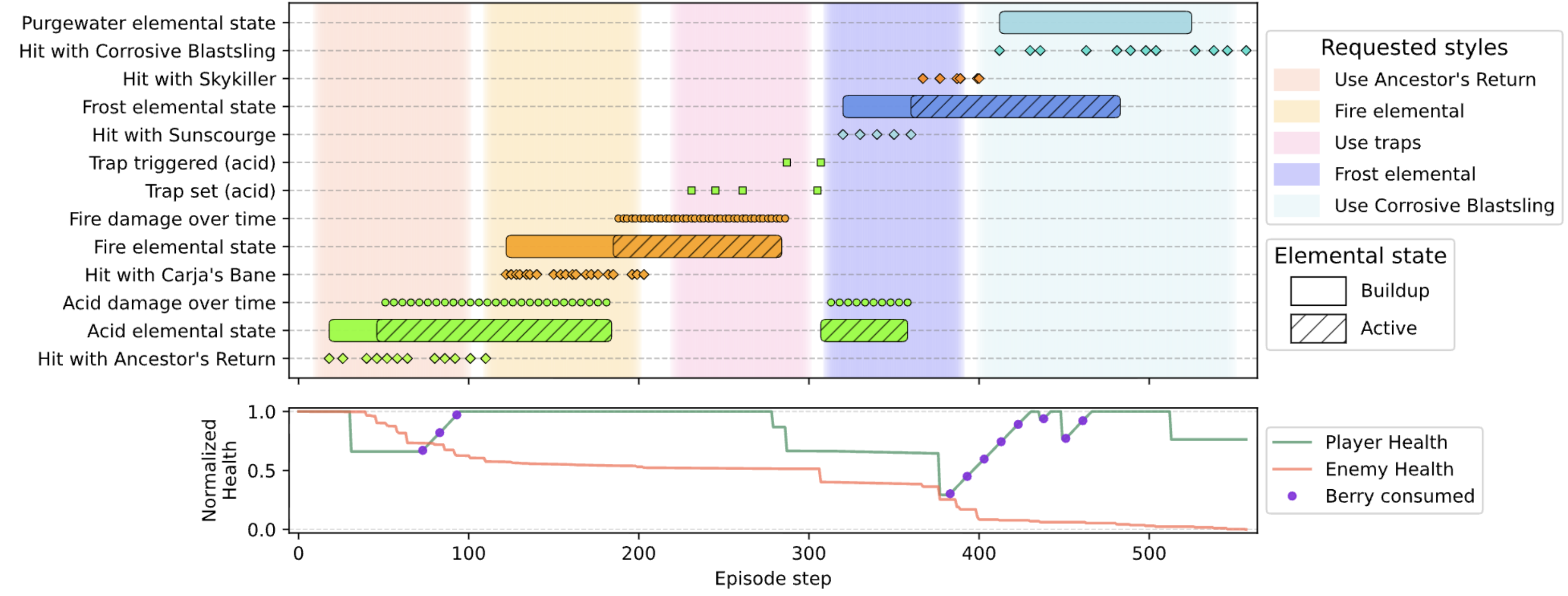}
\caption{
\textbf{Timeline:} The timeline of one selected battle against an Apex Thunderjaw in which the agent is scripted to switch through a sequence of four styles, as if it were being coached at execution time.
For elemental styles, the agent is rewarded only for the damage done while the machine is in the desired state (hashed regions).
The agent uses one weapon-ammo combination to achieve the state but then, depending on the power of the weapon in hand, may switch to a more powerful weapon to maximize damage while the state lasts.
The image also shows the normalized health of Aloy and the machine, and how she consumes berries to recover health.
}
\label{fig:timeline_breakdown}
\end{figure}

While at training time we keep the requested style fixed throughout an episode, at runtime we can switch between style requests at any moment. Figure \ref{fig:timeline_breakdown} shows what that looks like in practice. For a single episode of Aloy fighting a Thunderjaw, we switch between five different styles and show how this affects various runtime metrics.

We also performed a more systematic study of switching statistics for the 6 weapon styles. In particular, we computed the number of steps it takes from the moment we change the requested style from weapon-type-1 to weapon-type-2, for all possible weapon pairs. The number of steps averaged across all weapon-pairs and enemies is 3.5 (with 5 actions/steps per second). We also computed the average win rate and style score, which is close to 1 for both. These results show that style switching is responsive (in under one second of gameplay), does not degrade task performance, and works across all style combinations tested. The full results can be found in Figures \ref{fig:S2S switching statistics} and \ref{fig:S2S damage matrix} of Appendix \ref{app: extended data}.

\subsubsection{Out-of-Distribution Performance}

The HFW training scenarios consist of Aloy facing a single enemy, drawn from a set of 19 different enemies. Furthermore, the style-parameter vector that is sampled is almost exclusively one for which all style-weights except one have a value of 0 (in other words, only 1 style is activated). Given these training settings, we explored 3 types of out-of-distribution performance: 1) fighting against unseen enemies, 2) fighting against two enemies at the same time, and 3) activating two styles at once.

Table \ref{tab:OOD scenarios} shows performance when Aloy fights against enemies not in its training set (the Stormbird, the Tremortusk and the Dreadwing), and when it fights a pair of its toughest enemies (2 Thunderjaws or 2 Specter Primes). Overall, the agent shows good style and task performance against enemies it wasn't trained on. In particular, its performance against the Dreadwing, a flying machine, is impressive, given that none of the 19 enemies from its training set can fly. Generalization to 2 enemies also works reasonably well. Obviously, win rate goes down as these new scenarios are more challenging, but the agent continues to perform competitively.

\begin{wraptable}{r}{0.43\textwidth}
\centering
\caption{Five OOD scenarios compared with two in-distribution scenarios.}
\label{tab:OOD scenarios}
\small
\setlength{\tabcolsep}{6pt}
\renewcommand{\arraystretch}{1.15}
\begin{tabular}{@{} l l l @{}}
\toprule
\textbf{Enemy} & \textbf{Style score} & \textbf{Win rate} \\
\midrule
Thunderjaw & 0.77 & 0.75 \\
Specter Prime & 0.79 & 0.93 \\
\midrule
2 Thunderjaws & 0.69 & 0.23 \\
2 Specter Primes  & 0.74 & 0.33 \\
Stormbird & 0.71 & 0.89 \\
Tremortusk & 0.76 & 0.88 \\
Dreadwing & 0.77 & 0.95 \\
\bottomrule
\end{tabular}
\end{wraptable}
We also evaluated dual-style activation, which represents a different type of OOD scenario. During training, we sample UVFA parameters independently 10\% of the time, making dual-style activation statistically negligible ($p \approx 10^{-16}$ to sample, for 16 out of the 18 style weights, a value within a near-zero interval that makes up 10\% of the full sampling range). 

Conflicting style requests naturally cause the agent to favor whichever style maximizes the return. However, combining a weapon or trap style with an elemental ammo style is a natural, non-conflicting option. We tested each of the 16 valid combinations (a particular weapon or trap typically has access to only 1 to 3 elemental ammo types). Full results are shown in Figure \ref{fig:HFW-dual-style-damage-done} of the Appendix. Overall, the agent demonstrated good OOD performance: the requested weapon was dominant in 15 of 16 cases, and the requested elemental ammo was dominant in 12. And whenever a requested weapon or ammo fell short of top dominance, it consistently placed second.

\subsubsection{Ablations}
\label{sec:ablations}

\begin{wrapfigure}{r}{0.5\textwidth}
    \centering
    \includegraphics[width=\linewidth]{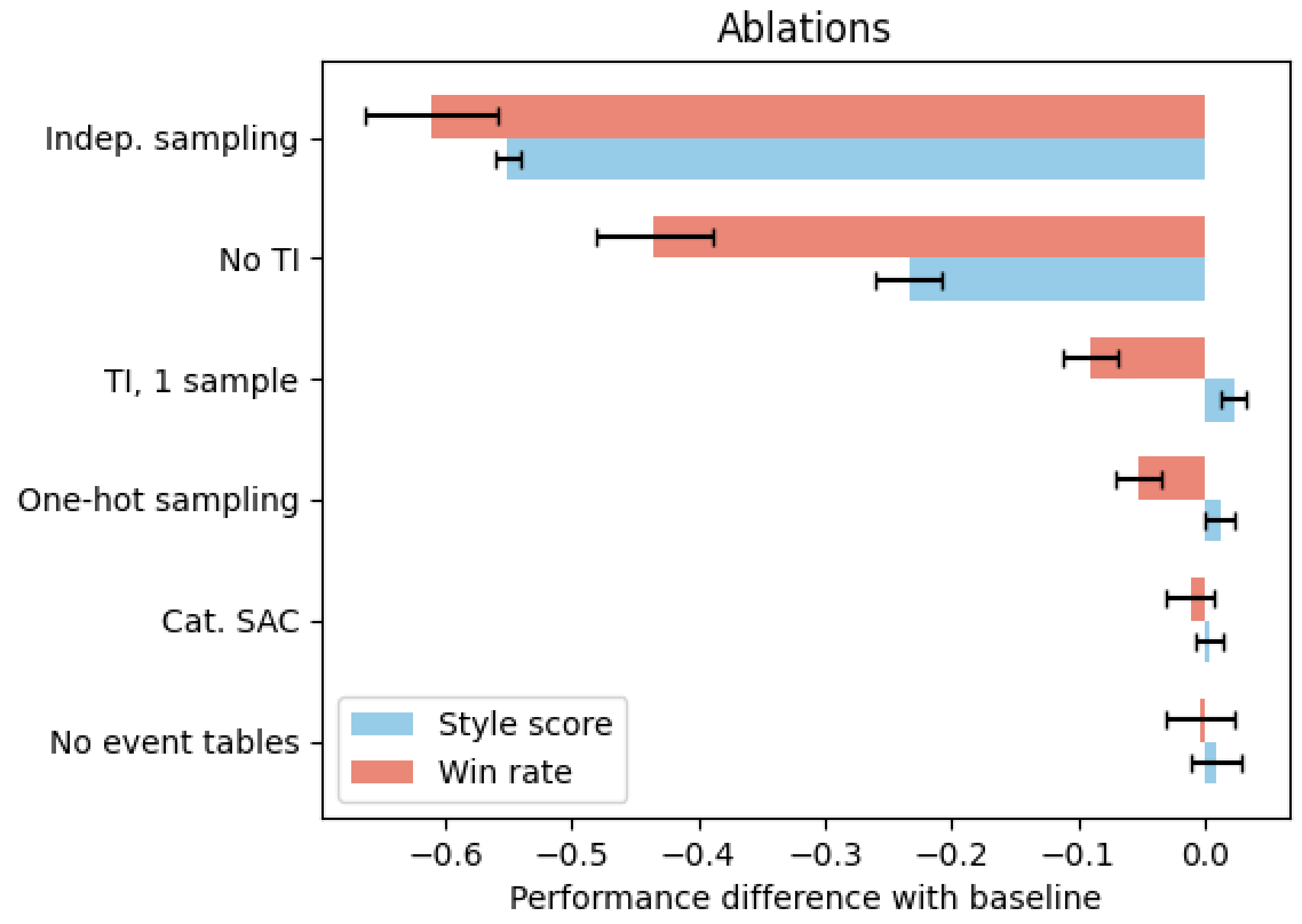}
    \caption{\textbf{Ablation study: }Performance difference relative to the baseline approach, for win rate and style score. TI refers to Task impersonation.}
    \label{fig:ablations}
\end{wrapfigure}

Figure \ref{fig:ablations} shows the main results of an ablation study that we did to investigate the impact of various framework choices.  Bars report the performance difference from the baseline setup against the two most difficult enemies: the Apex
Thunderjaw and the Slaughterspine. All bars represent the average across five training runs with the full range of the samples displayed as an error bar. 

As mentioned, the baseline style-weight sampling strategy mainly consists of continuous one-hot sampling: all style-weights are 0, except for one, whose value is sampled uniformly at random from its corresponding range. If we replace this dominant sampling-strategy with independent sampling (all style-weights sampled uniformly at random from their ranges independently), performance drops significantly. This is not surprising, given that our evaluation scenarios only consider a single active style. Having many (potentially opposing) styles active at the same time creates a less useful learning signal. What is surprising is that regular one-hot sampling, which sets the active style to a preselected value that is also used to evaluate that style, is also worse. With more diverse training scenarios, the continuous one-hot sampling strategy results in overall stronger performance.

Besides style-weight sampling strategies, the use of task impersonation (TI) is critical. Both style performance and win rate drop significantly without it. The baseline uses TI with 4 impersonation samples, which strikes a good balance between performance and wall-clock time. Using TI with 1 sample leads to a drop in win rate, while it is still better than no TI at all.

We also did an ablation that changed the base algorithm of our baseline from categorical RAC to categorical SAC (categorical critic with classical SAC updates), resulting in a tiny reduction in win rate. Finally, while stratified sampling was important for GT, the event tables used in HFW turned out to play a negligible role. 

\subsection{Humanoid}

For the humanoid domain, we evaluate three arm styles and two gait styles (short and long) at fixed intervals during training. We say an arm style is successful if the arm angles are within a small window of the arm-angles corresponding to the requested arm style; a gait style is successful if the gait is within a small window of the requested gait. The training plots are shown in Figure \ref{fig:humanoid results} in Appendix \ref{app: extended data}. The results show that for the reward-conditioned UVFA method, after training for about 1000 epochs, the arm-style and gait-style are successful in about 95\% of the cases. By contrast, for a non-UVFA baseline the arm-style is never successful and the gait-style only in about 5\% of the cases. The required
task performance (obtaining a forward velocity of at least 1) is achieved already at the first evaluation point at epoch 50, for all methods.

\section{Conclusions}

Coachable agents have immediate uses in gaming, robotics, and other fields where RL is used to solve real-time control problems.
Style-conditioned UVFA, with the appropriate learning algorithms, scenario training, and sampling strategies, can produce agents that can succeed at complex tasks---like walking, driving, or fighting in a video game--while being able to modify their style on command.
AI systems trained in this manner have broad practical applications because, rather than learning a single behavior, they encode a family of behaviors.

The technology can have an immediate impact on the \$300B video game industry by improving QA processes or enhancing in-game NPCs.
It can also lead to novel accessibility features to enable a broader audience to enjoy video games by letting agents take over when the game is too difficult.
The advanced interactivity these agents provide can open up new gaming experiences for players and new design options for game developers.

Outside of games, as more general purpose robots make their way out of labs, it is inevitable that their human users will have varying preferences over how the machines perform their tasks. 
The approaches described here offer a powerful mechanism to enable users to shape a robot's behaviors with goal modifiers. 
Perhaps soon we will have robots that can not only clean for us, but that will also be able to do so with flair.

\subsubsection*{Author Contributions}
R.C. and H.v.S. managed the project and led the research and development efforts across domains.
N.B., J.D., D.M., D.O., and F.R. supervised parts of the research and development efforts for the HFW \playstyles.
I.D., S.G., R.K., E.A., and A.L. participated in the research and development of the HFW agent \playstyles.
A.S., K.N.F., J.G., V.R.K., J.A.M., and K.W. participated in the research and development of the HFW agent.
T.J.W. led the research and development in DMC Humanoid.
N.D.B, D.H., and S.M. participated in the research and development of the DMC Humanoid agent.
S.J. curated the open source release of the algorithm and the DMC Humanoid experiments.
K.S. led the research and development for \GTSophy/.
A.D., Y.D., K.K., P.M., C.S., F.F., T.S. participated in the research and development of \GTSophy/.
F.D., G.Z.H., S.B., M.G., M.S., M.D.T., and J.B.T. provided engineering support and built the libraries and APIs that connected to the different videogames. 
P.S. and M.S. provided executive support and technical and research advice.
P.W. oversaw the project, provided executive support, resources, and technical advice and managed stakeholders. 

\subsubsection*{Acknowledgments}
We thank our collaborators at Guerrilla Games and Polyphony Digital who allow us to use their games for research. 
We also thank our colleagues in the platform team at Sony AI who build and maintain the infrastructure that supports our experimentation.

\clearpage

\bibliography{main}
\bibliographystyle{tmlr}

\clearpage
\appendix
\section*{Appendices Overview}

\begin{figure}[ht]
    \centering
    \includegraphics[width=0.32\textwidth]{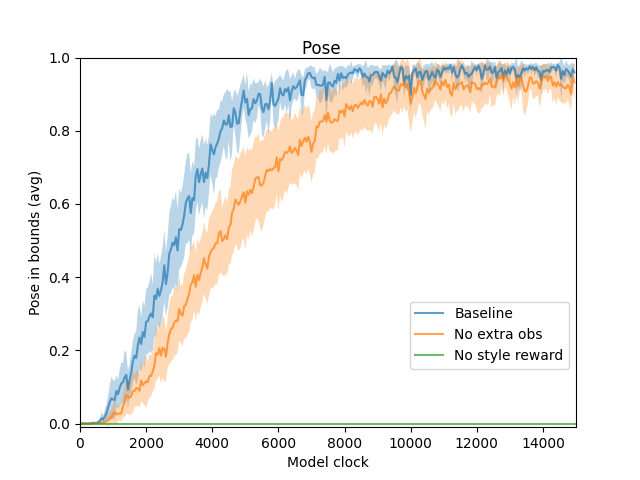}
    \includegraphics[width=0.32\textwidth]{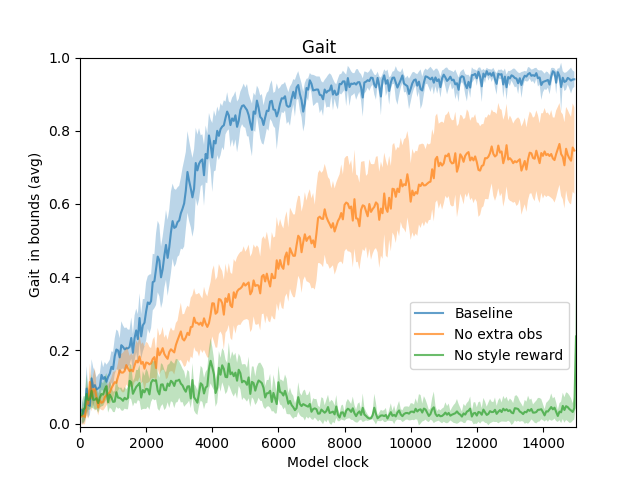}
    \includegraphics[width=0.32\textwidth]{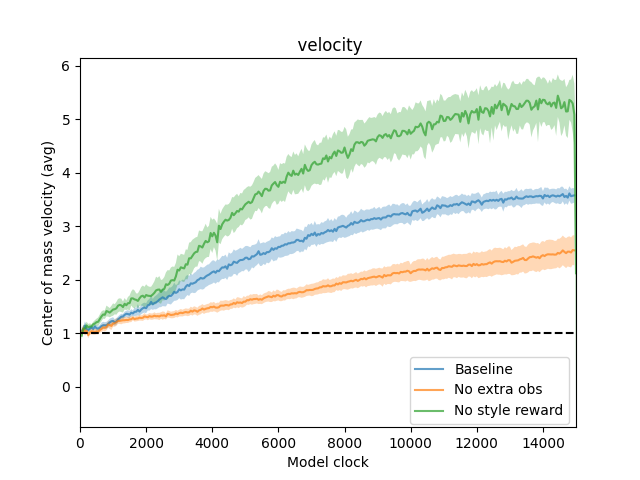}
    \caption{\textbf{Humanoid Domain Results.}
Ablation of agent performance on styles and base walk task. All results were averaged across 6 evaluated style combinations (3 arm styles and 2 leg styles) based on 20 training runs. Shaded regions indicate 95\% confidence intervals. Style performance is shown as the percentage of style constraints achieved against epochs (learning curves of style) for arm poses (\textit{left}) and leg gait (\textit{middle}). Task performance requires walking only at a speed of at least 1, which all of the agents achieve at the first evaluation point already (at model clock 50). The ablation without a style reward moves faster since it does not need to adhere to a particular gait or arm pose (\textit{right}). Ablations examine agent performance when removing additional observation features relevant to the leg style reward, and when the agent is trained without style rewards.
    }
    \label{fig:humanoid results}
\end{figure}

\begin{figure}[ht]
\centering
\includegraphics[width=\linewidth]{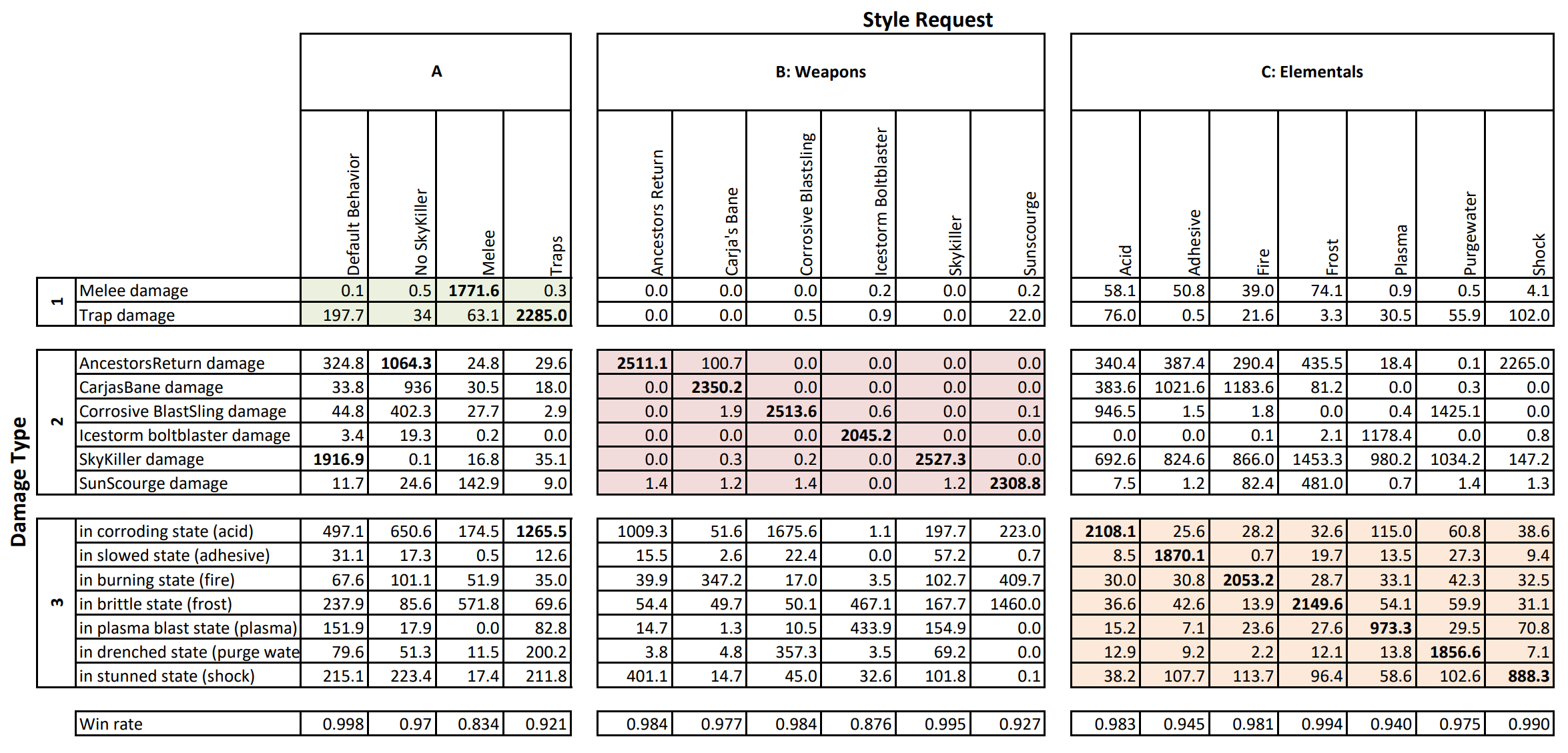}
\caption{
\textbf{HFW Style Damage Done:}
This matrix shows the agent's overall ability to meet the style objectives. 
The columns correspond to the 17 styles measured by damage inflicted, and the rows are damage metrics, with a max value of 2747.25 (the average of all of the machines' health).
Note that the aggregation of damage across the different categories will not always sum to the same value; for example, when asked to use Ancestor's Return weapon, the agent frequently uses the acid ammo, and the damage inflicted counts towards both the weapon and the elemental damage type.
The data is collected from fights against all \NumMachinesInUse\ enemies 10 times in each of the three test locations (a total of 2850 battles per style).
}
\label{fig:HFW-damage-done-by-style}
\end{figure}

\begin{figure}[ht]
\centering
\includegraphics[width=\linewidth]{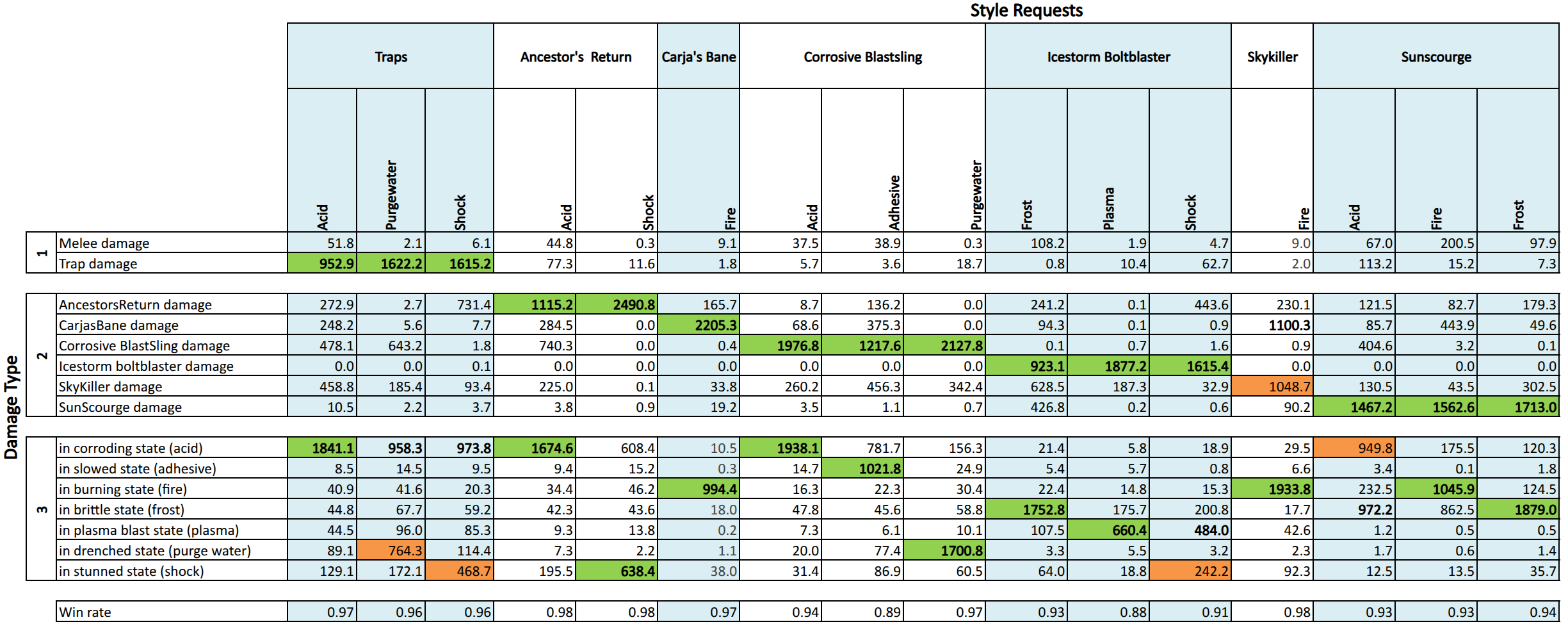}
\caption{
\textbf{HFW Dual-Style Damage Done.} 
This matrix shows the damage-done statistics when a weapon or trap style is requested together with an elemental-ammo style, yielding a total of 16 valid style combinations. The highlighted cells are the damage-type cells corresponding with the style requests. Green means the requested style is the dominant one (i.e., yields the highest damage within its group); orange means the requested style is not dominant. In other words, for full compliance with the style requests, all highlighted cells should be green. 
}
\label{fig:HFW-dual-style-damage-done}
\end{figure}

\begin{figure}[ht]
    \centering
    \includegraphics[width=0.9\textwidth]{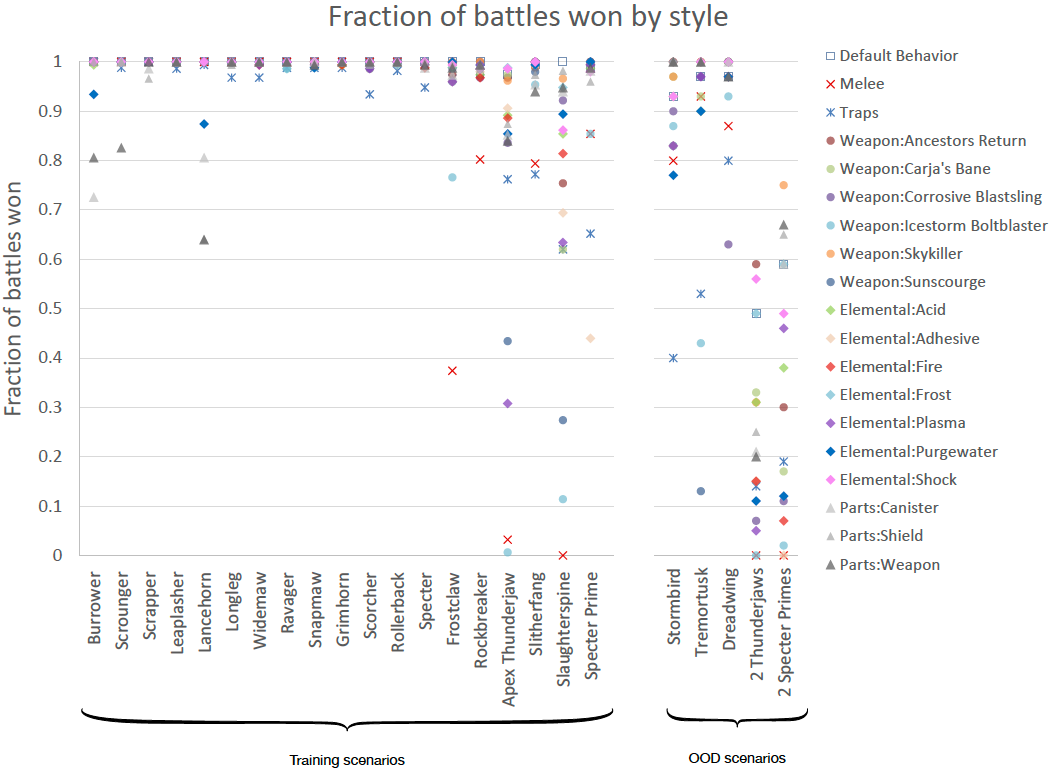}
    \caption{\textbf{Fractions won by style:}
    For each style, the fraction of battles that the agent wins against
each type of enemy. These results highlight the styles that are less effective, particularly against the
tougher enemies, including melee, the Boltblaster and Sun Scourge, and some elementals. In the five OOD scenarios to
the right, including three machines not in the training set, the agent also does quite well. Not unexpectedly, performance
drops in the rightmost OOD scenarios where the agent faces a pair of the toughest enemies (a scenario not seen in the
actual game), though the agent still puts up a good fight.
    }
    \label{fig:fractions_won_by_style}
\end{figure}

\begin{figure}[htbp]
  \centering
  
  \includegraphics[width=0.77\textwidth]{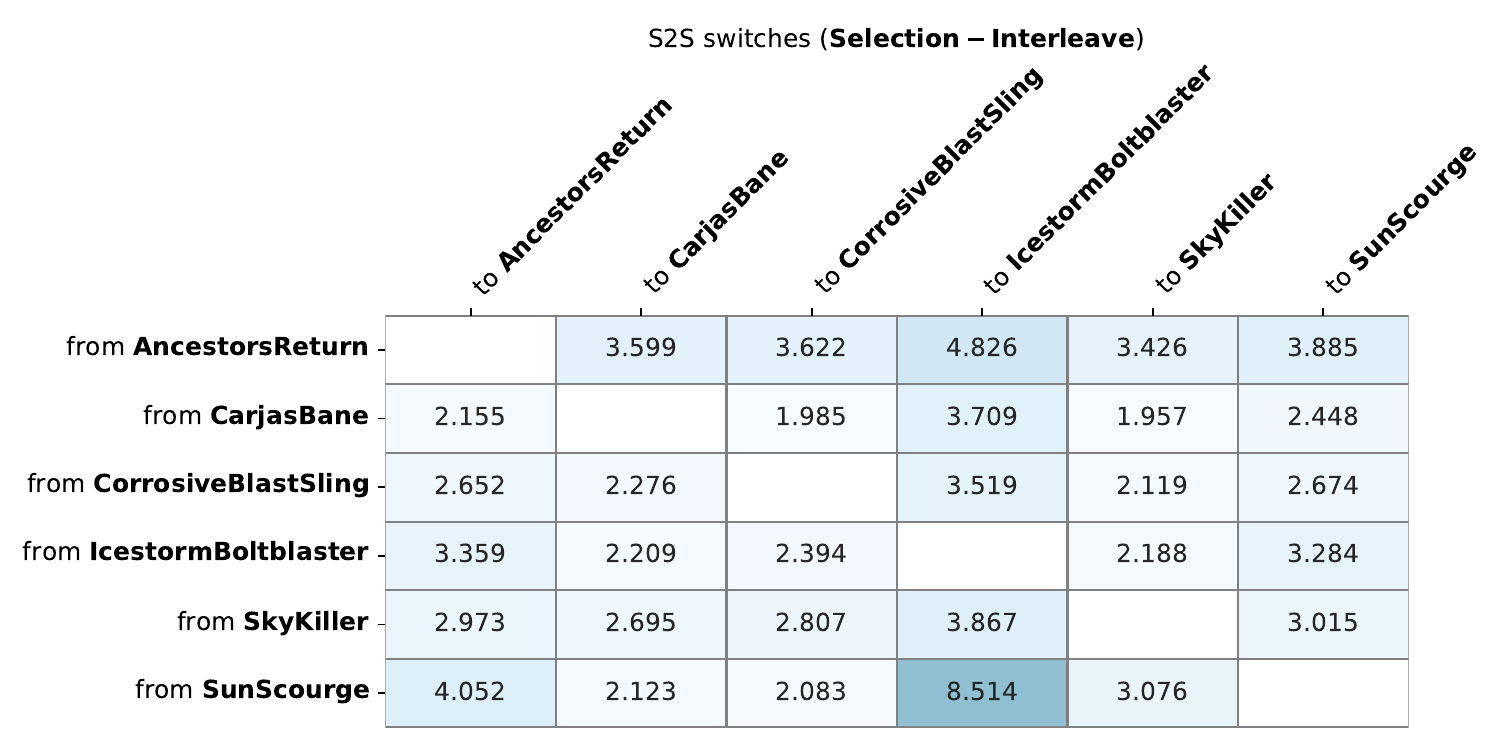}
  
  \vspace{1em} 
  
  \includegraphics[width=0.77\textwidth]{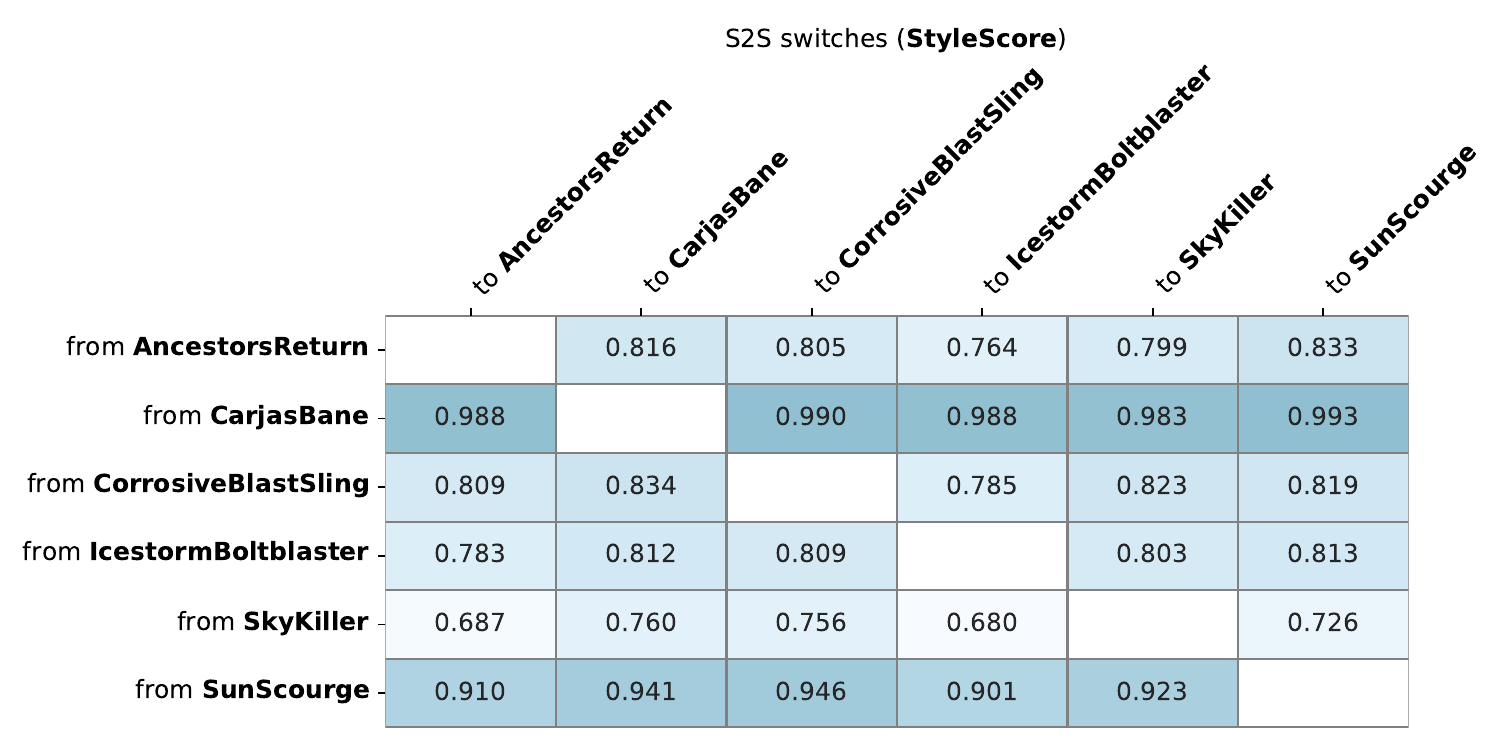}
  
  \vspace{1em}
  
  \includegraphics[width=0.77\textwidth]{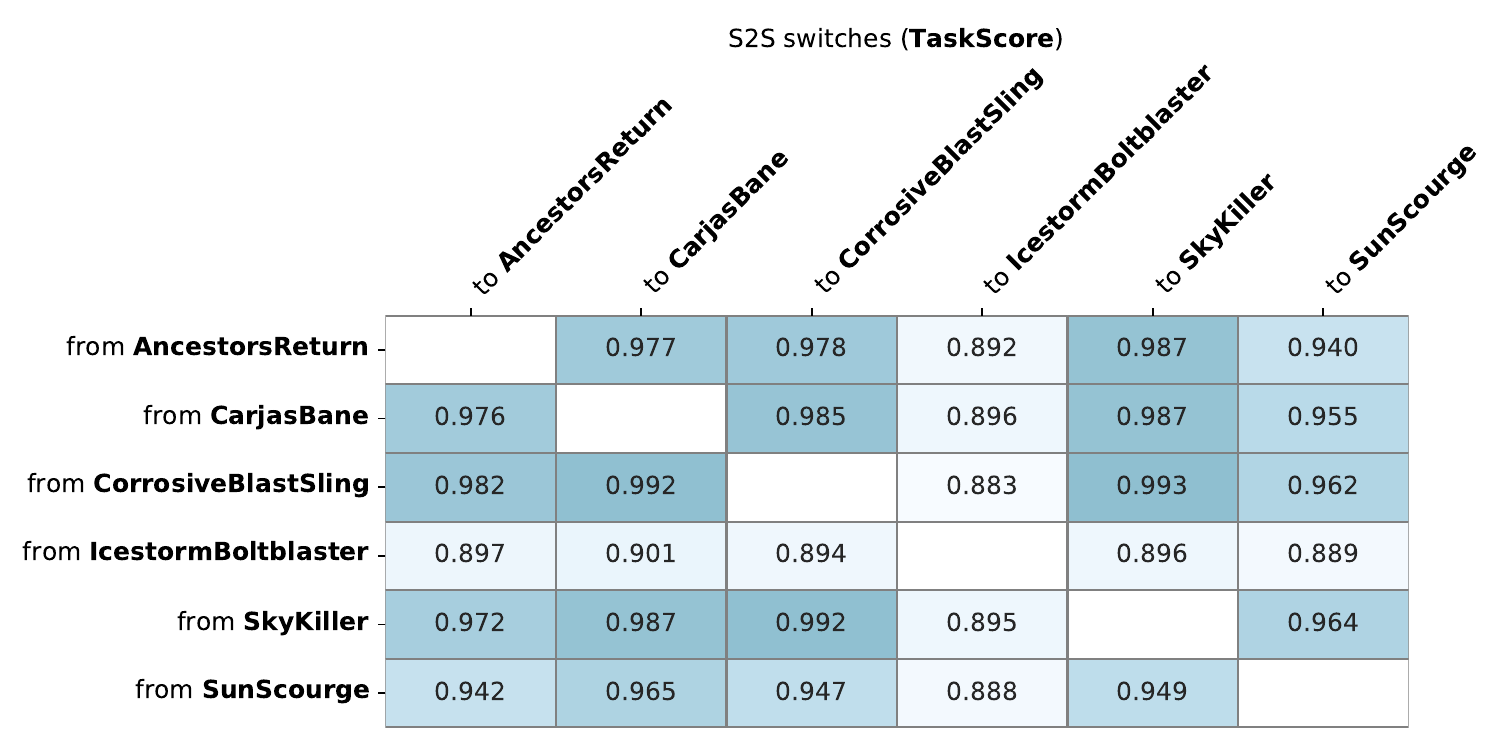}
  \caption{\textbf{HFW Style-to-Style Switching Statistics.} Results from switching from \textit{weapon-type-1} to \textit{weapon-type-2} the moment enemy-health drops below 50\%, averaged across all enemies and 5 independent training runs. \textbf{Selection-Interleave} indicates the `step in which weapon-type-2 is first selected' minus the `step in which the style-request switched'. \textbf{StyleScore} indicates the `damage dealt by weapon-type-2' normalized by the `total damage dealt after the style-switch'. \textbf{TaskScore} is equal to the win rate. }
  \label{fig:S2S switching statistics}
  
\end{figure}

\begin{figure}[ht]
    \centering
    \includegraphics[width=1.0\textwidth]{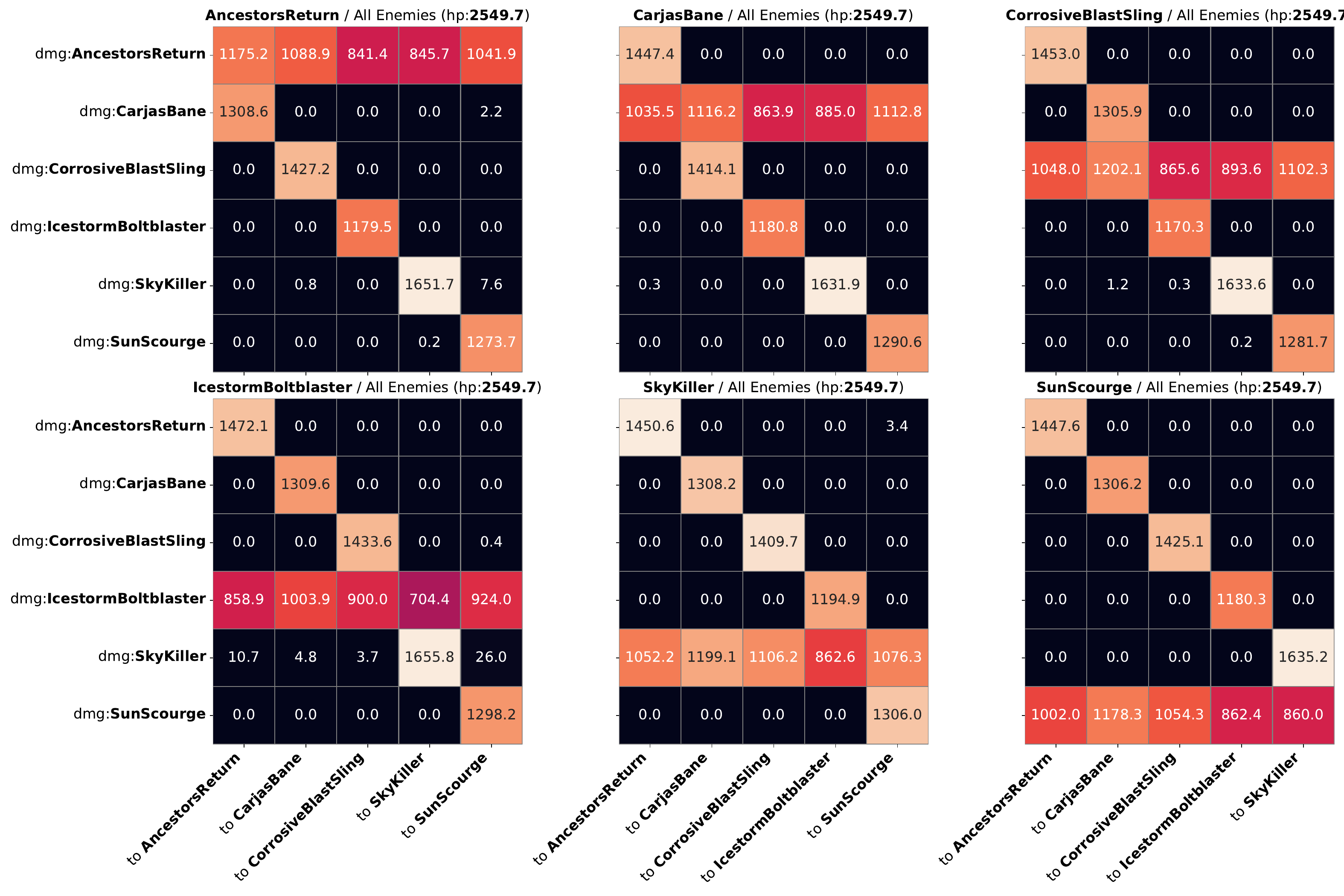}
    \caption{\textbf{HFW Style-to-Style Damage Matrices.}
Damage matrices when switching from \textit{weapon-type-1} to \textit{weapon-type-2} once enemy-health is at 50\%, averaged across all enemies and 5 independent training runs. The matrices show that the damage done to an enemy is roughly split between the two requested weapons, as expected.}
    \label{fig:S2S damage matrix}
\end{figure}

\begin{figure}[ht]
\centering
\includegraphics[width=\linewidth]{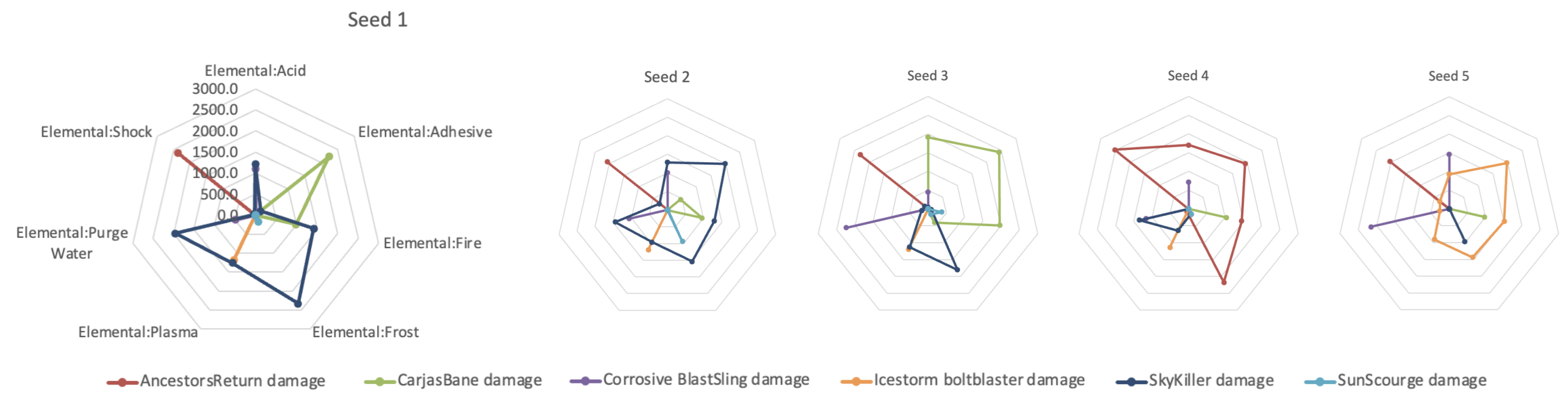}
\caption{
\textbf{Seed Comparison:} These five radar plots show how much damage was inflicted by each weapon for each of the seven elemental styles for each of the five training seeds used to generate Figure \ref{fig:HFW_style_vs_damage}.
These five charts show that each of the five policies favors different combinations of weapons and ammo once the machine is in an elemental state which, when averaged across, leads to the speckled pattern in block C2 in Figure \ref{fig:HFW_style_vs_damage}.
}
\label{fig:seed_breakdown}
\end{figure}

\begin{itemize}
    \item Appendix \ref{app:style-conditioned UVFA}: Style-Conditioned UVFA
    \item Appendix \ref{app: training algorithms}: Training Algorithms
    \item Appendix \ref{app: domain details}: Domain Details
    \item Appendix \ref{app: extended data}: Extended Data 
    \begin{itemize}[labelwidth=1.8cm, leftmargin=2.0cm, align=left]
        \item[$\circ$~Table~\ref{tab:domain_training_summary}]- Per-domain training configuration.
         \item[$\circ$~Figure~\ref{fig:humanoid results}]- Humanoid Domain Results.
        \item[$\circ$~Figure~\ref{fig:lat-control}]- GT: aggressiveness control.
        \item[$\circ$~Table~\ref{tab:machineDetails}]- HFW: machine types reference.
        \item[$\circ$~Table~\ref{tab:weaponsDetails}]- HFW: configured weapons reference.
        \item[$\circ$~Table~\ref{tab:enemyRecords}]- HFW: win-loss records.
        \item[$\circ$~Figure~\ref{fig:HFW-damage-done-by-style}]- HFW: Damage by style, numbers.
        \item[$\circ$~Table~\ref{tab:baseline style score win rate}]- HFW: style scores and win rates.
        \item[$\circ$~Figure~\ref{fig:HFW-dual-style-damage-done}]- HFW: Damage by dual styles.
        \item[$\circ$~Figure~\ref{fig:fractions_won_by_style}]- HFW: Fractions won by style.
        \item[$\circ$~Figure~\ref{fig:S2S switching statistics}]- HFW: style-to-style switching statistics.
        \item[$\circ$~Figure~\ref{fig:S2S damage matrix}]- HFW: style-to-style damage matrices.
        \item[$\circ$~Figure~\ref{fig:seed_breakdown}]- HFW: seed comparison.
        \item[$\circ$~Table~\ref{tab:all_rewards_descriptions_hfw}]- HFW: reward descriptions.
        \item[$\circ$~Table~\ref{tab:static_rewards_hfw}]- HFW: task rewards.
        \item[$\circ$~Table~\ref{tab:dynamic_rewards_hfw}]- HFW: style rewards.
        \item[$\circ$~Table~\ref{tab:hfw_policy_heads}]- HFW: policy head outputs.
        \item[$\circ$~Table~~\ref{tab:feat_rankings_two_methods}]- HFW: Feature importance analysis.
    \end{itemize}
    \item Appendix \ref{app: videos}: Videos
\end{itemize}

\clearpage
\section{Style-Conditioned UVFA}\label{sec:trainingAlg}
\label{app:style-conditioned UVFA}

We train our policies using universal value function approximation (UVFA)~\citep{pmlr-v37-schaul15} within an actor-critic framework.
Classical UVFA conditions the policy, value, and reward functions on both the state and a goal $g$, producing goal-dependent behavior. Our \emph{style-conditioned UVFA} instead fixes the core task (e.g., defeat the enemy) and exposes runtime style parameters $\Theta$ that specify \emph{how} the task is pursued (e.g., weapon preference). Both policy and critic are conditioned on $\Theta$, letting a single policy express multiple behavioral modes without retraining.
Below, we formalize the style-parameter interface $\Theta$, its encoding, and the parameter-sampling strategies used during training.

\paragraph{Formalization.}
We encode the active style via $f(\theta_t)$ and form the augmented state $\tilde{s}_t\coloneqq(s_t,f(\theta_t))$.
The policy and critic then condition on $\tilde{s}_t$ as $\pi(a_t\mid\tilde{s}_t)$ and $Q(\tilde{s}_t,a_t)$.
With discount $\gamma\in[0,1)$ and horizon $T$, the per-step reward is 
\[
r_t=r_{task}(s_t,a_t)+ r_{styles}(s_t,a_t;\theta_t),
\]
where $r^{task}$ is the task reward (fixed across time/episodes), and $r^{style}$ is the style reward that depends on $\theta_t\in\Theta$. Typical style parameters include weights, thresholds (for gating/ramping), and targets (setpoints) that penalize deviation from desired ranges.

\noindent In the general case, these style parameters may evolve  within an episode as a sequence $\theta_{0:T-1}$ with
\[
p_{\Theta}(\theta_{0:T-1})=p_{\mathrm{init}}(\theta_0)\,\prod_{t=0}^{T-2}\kappa(\theta_{t+1}\mid \mathcal{C}_t),
\]
where $p_{\mathrm{init}}$ is the initial distribution and $\kappa$ is a (generally unknown) transition rule that may depend on context \(\mathcal{C}_t\) (e.g., user input, schedules, or other signals). This style-parameter dynamics induces the augmented transition $P(s_{t+1}\mid s_t,a_t)\kappa(\theta_{t+1}\mid\mathcal{C}_t)$.
However, since \(\kappa\) is unknown at training time, we fix \(\theta_t\) within each training episode, thus avoiding credit assignment through \(\kappa\). In our studies, this approach empirically accelerated learning while preserving test-time controllability via \(\theta\). 
Given $\theta$ sampled at episode start, the return and objective are
\[
G_{\theta}\;=\;\sum_{t=0}^{T-1}\gamma^{~t}\!\Big[r_{task}(s_t,a_t)+ r_{styles}(s_t,a_t;\theta_t)\Big],\qquad
J(\phi)\;=\;\mathbb{E}_{\substack{s_0\sim \rho,\, \theta\sim p_{\Theta} \\ \tau \sim \pi_\phi}}\!\big[\,G_{\theta}\,\big],
\]
where $\rho$ is the initial-state distribution, $p_{\Theta}$ is the episode-level sampling distribution over styles, and $\tau$ is the trajectory induced by $\pi_{\phi}$ under $\theta$.

\paragraph{Implications.}
Conditioning the policy on style parameters $\Theta$ while keeping the core objective $r_{task}$ fixed allows one policy to express many behaviors without retraining.  This methodology replaces multiple specialized policies, lowers training cost, and enables systematic performance-vs-style trade-off sweeps over $\theta\in\Theta$. Encoding styles on distinct parameters enables a compositional space of preferences: they can be turned on independently, combined, or varied continuously. When a request conflicts with the core task, the resulting behavior reflects the relative weighting of the task and style terms: the agent trades off task return against style adherence according to the configured reward scales.
Styles encode designer preferences, 
so their rewards must be carefully designed. Agents may learn well for one $\theta$ yet generalize poorly to $\theta'$. 
Because we condition the policy and value function on $\theta\in\Theta$, stable learning benefits from a well-behaved dependence on such parameters: nearby style settings should induce comparable return targets, which we encourage by bounding/normalizing style-reward magnitudes. Accordingly, data sampling should cover the style settings expected at evaluation time or deployment.
At execution, $\Theta$ enables real-time behavioral modulation, reliable only when style axes are interpretable and parameters stay within trained regions. 
Within these limits, the agent responds smoothly to changes in $\theta$. 
Evaluation must therefore cover representative regions of $\Theta$ to verify coherent, responsive stylistic control.

\paragraph{Parameter Sampling.}
As defined above, the reward decomposes into a task term and a style term.
Previous work explored related ideas in small, turn-based games via configurable reward shaping~\citep{de2021configurable,de2022automated}, typically using linear weight vectors $w$. Building on that formulation, we apply style-conditioned rewards in complex, high-dimensional, real-time control domains by decomposing the style reward into $m$ distinct style terms:
\[
r_{styles}(s_t,a_t;\theta) \;=\; \sum_{i=1}^m w_{i}\; g_i(s_t,a_t;\xi_{i}),\qquad \theta=(w_0, \dots, w_m, \xi_{0}, \dots, \xi_{m})
\]
where $w_{i}\in\mathbb{R}$ is a scalar weight, $g_i$ a style reward, and $\xi_{i}$ a (possibly zero-dimensional) style parameter vector. Both $w_{1:m}$ and $\xi_{1:m}$ are part of the observable style vector $\theta$ (hence in $f(\theta)$). By convention, $w_i>0$ encourages style $i$, $w_i<0$ discourages it, and $w_i=0$ expresses no preference.

During training we sample each $w_i$ from the parameter's domain. In some environments, it is helpful to extend the training domain beyond the range used at evaluation time. For example, in GT we expand $[0,k]$ ranges to train on $[-i,0,j]$, where $i\geq 0$ and $j\geq k$, to encourage the agent to learn a spectrum of behaviors, rather than collapsing to an ``average'' policy that partially satisfies both 0 and positive values. Using 0 to indicate ``not encouraged'' (as opposed to negative values meaning ``discouraged'') also enables more coherent behavior when multiple, overlapping styles are combined.

For instance, encouraging part targeting without pushing the agent away from general weapon use is straightforward if $w_{\text{part}}>0$ while $w_{\text{weapon}}=0$, whereas forcing only $\{\text{positive},\text{negative}\}$ on both axes induces undesirable trade-offs.

Domains may expose discrete or continuous style knobs: categorical choices (e.g., a selector) are naturally discrete, while weights are continuous. In practice, we found that continuous weights appear to be as easy to learn as discretized sets (e.g., $\{a,0,b\}$) and 
preserve post-training reward tuning without retraining.

\clearpage
\section{Training Algorithms}
\label{app: training algorithms}

Because our experimental domains differ in complexity, dynamics, and action spaces, we use different RL algorithms per environment (Table~\ref{tab:domain_training_summary}, Appendix \ref{app: extended data}) while keeping a common off-policy actor-critic backbone with replay.
For all domains, we use
many asynchronous rollout workers that pull the latest policy networks and sample UVFA weights at episode start (fixed within an episode), collect data and stream transitions to a shared replay buffer. A single learning procedure updates the policy/critic, details of which are discussed below. Episodes end on success/failure (e.g., all opponents defeated) or on a fixed time limit. Style conditioning follows the UVFA formulation above. \textbf{Scenario training}\citep{li2022metadrive,cobbe2020leveraging,tobin2017domain} is used in all domains to expose the agent to diverse initializations and task variants. \textbf{Stratified sampling}, in the form of multi-tables\citep{wurman2022outracing} or event tables\citep{kompella2023eventtables},
enables replay by scenario/style to balance mini-batches, ensuring that multiple situations are represented in each network update. An adaptation of \textbf{task impersonation}~\citep{kalashnikov2021task-impersonation}
augments each sampled mini-batch by relabeling transitions under additional UVFA parameter vectors drawn from the same distribution, creating $(n{+}1)$ labeled versions of each sampled transition at update time without writing duplicates to replay.

Next, we summarize the different training algorithms that we use: SAC for the Humanoid domain, QR-SAC for Gran Turismo and (Cat-)RAC for Horizon. Unless noted otherwise, we optimize with Adam (TensorFlow defaults; learning rates and other hyperparameters are reported with the experiments) and use target networks for stability.

\subsection*{SAC}

For the Humanoid domain, we use Soft Actor–Critic (SAC)~\citep{haarnoja2018soft}, that maximizes an entropy-regularized (``soft'') return by learning a stochastic policy and twin critics. The critic targets use TD(0) bootstrapping with a \emph{minimum} over two Q-networks to reduce overestimation bias, and slowly updated target networks (Polyak averaging). The policy update is based on the critic's gradient; for continuous actions we use the reparameterization trick for $\tanh$-Gaussian heads so gradients flow from $Q$ through the sampled action to the policy parameters. Using this approach, a sample $X$ from a $\gaussian(\gaussianMean, \gaussianStd^2)$, is computed as $X = \gaussianMean + \gaussianStd\cdot Z$, where $Z \sim \gaussian(0, 1)$.
In our setup, we follow the canonical SAC formulation and perform \emph{simultaneous} actor/critic updates each step using batches from the replay buffer, while the entropy temperature is fixed.

\subsection*{QR-SAC}
For \GranTurismo7/ we adopt the quantile-regression variant of SAC (QR-SAC) from \cite{wurman2022outracing}, replacing scalar Q with a distributional critic trained by quantile regression while keeping the SAC structure (stochastic policy, replay, target networks). In this distributional setting we update the policy \emph{before} the critic each step, which we found to stabilize the moving value distribution.

\subsection*{RAC and Cat-RAC}
 Horizon's \PlayStation5/ control mixes discrete button combinations with continuous sticks/triggers. To support mixed action spaces of this kind we generalize SAC to a \emph{regularized actor–critic} (RAC) objective that replaces the entropy term with a generic regularizer $\mathcal{R}(a\mid s,f(\theta))$. This modification lets us apply different regularizers per action subspace and sum them in a single objective while retaining the SAC training loop. Specifically, we apply \emph{exact} categorical entropy to the discrete policy heads and an \emph{n}-sample empirical mean of the negative log density from the $\tanh$-Gaussian policy heads. In our experiments, we estimate the negative entropy by averaging $n=10^3$ samples per component, instead of the conventional single sample. This approach is substantially more accurate than the single-sample surrogate commonly used in SAC, without a meaningful slowdown in practice. Using exact categorical entropy is necessary because gradients cannot be propagated through categorical samples.

As in SAC, we use the reparameterization trick for the Gaussian policy heads. A reparameterization trick also exists for a ($d$-dimensional) categorical distribution, $\pi \in \simplex^d$: the Gumbel-max trick\citep{gumbel1954statistical} enables the computation of a sample $Y \sim \pi$ as
$Y = \argmax_{i} \log \pi_i + Z_i$,
where $Z = \set{Z_i \sim \mathrm{Gumbel}(0, 1)}_{i = 1}^d$.
However, the presence of the argmax operation prevents this reparameterization from being differentiable.
For this reason, we employ a Gumbel–Softmax continuous relaxation~\citep{jang2017categorical,maddison2017concrete} where we query the critic with relaxed actions $\exp\!\big(\log \pi + Z\big)\,\big/\,\sum_{i=1}^d \exp\!\big(\log \pi_i + Z_i\big)$ to compute critic gradients w.r.t.\ categorical policy heads during policy updates, while the environment executes ``hard'' action samples\footnote{%
See \citet{huijben2023reviewOfGumbelMaxTrick} for a survey of Gumbel-max/Gumbel-Softmax methods.}. A naive ``softmax over logits'' can be numerically brittle in our setup (temperature sensitivity, overflow errors). We therefore normalize the positive part of the network outputs $x\in\mathbb{R}^d$:
\[
\pi_i \;=\; \frac{\max\{0,x_i\}+\varepsilon}{\sum_j \big(\max\{0,x_j\}+\varepsilon\big)}, \qquad \varepsilon=10^{-8}/d,
\]
which stabilizes entropy and avoids pathological scaling.

To backpropagate through categorical heads, we use a regression regret-matching~\citep{waugh2015solving} update. Let $g=\nabla_{x}\mathcal{L}\in\mathbb{R}^d$ be the $d$-dimensional gradient of the policy loss w.r.t.\ pre-normalized outputs, and define the instantaneous regret
\[ \rho_i \;=\; \sum_{j} g_j\,\pi_j \;-\; g_i,\qquad i=1,\ldots,d. \]
We pass the modified gradient $g'$ to the optimizer, composed as
\[
g'_i \;=\;
\begin{cases}
0, & x_i<0 \ \text{and}\ \rho_i<0,\\
-\rho_i, & \text{otherwise.}
\end{cases}
\]
All other backpropagation is unchanged. Intuitively, this modified gradient shifts probability toward actions with positive regret while avoiding updates that merely push already-negative $x_i$ further negative, as only positive parts affect $\pi$. The modified gradient is very similar to the pseudoregret change in regret matching$^+$~\citep{cfrPlus,solvingHulhe} and, in our setting, yields stable, scale-robust gradients for categorical components without introducing temperature hyperparameters.
Regret matching~\citep{schmid2023sog,brown2019superhuman,brown2018superhuman,moravvcik2017deepstack,bowling2015hulheSolved,Hart00RegretMatching,Blackwell56RegretMatchingOrigin} provides theoretical guarantees about the optimality of its decisions that degrade gracefully as the capacity of its function approximator decreases~\citep{dorazio2019rcfr,morrill2022hrl,morrill2016rcfr}.

For representing values, we use a categorical distributional encoding of the critic on top of RAC, yielding \emph{Categorical RAC} (\catrac{}), inspired by \cite{farebrother2024stop}'s use of categorical value functions. The critic predicts a discrete value distribution; a fixed decoder (expectation under the support) produces scalars where needed. We train the critic with cross-entropy against HL-Gauss~\citep{imani2018improving} encodings of TD(0) targets (std/width ratio $0.75$, 255 bins, value range $(-600,1000)$). As with QR-SAC, we use a policy-first, then-critic update order to reduce non-stationarity in distributional targets.

\clearpage
\section{Domain Details}
\label{app: domain details}
As described in the main paper, we apply our approach to three domains that reflect our goals and history: \textbf{Horizon Forbidden West (HFW)}, \textbf{Gran Turismo~7 (GT)}, and \textbf{DMC Humanoid} from the DeepMind Control Suite.

\subsubsection*{Horizon Forbidden West}
\label{sec:hrz-domain-intro}
Large environments, complex interactions, and highly non-linear dynamics- together with long-horizon objectives and sparse/delayed rewards - make HFW a demanding RL testbed. 
A practical complication in such games is a cascade, or ``butterfly'' effect: small timing/state perturbations can cause two nearby trajectories to diverge within tens of frames, yielding different game-event graphs (e.g., failing to evade an attack, inaccurate weapon selection) and locally discontinuous returns.
We condition styles episodically, report aggregates over multiple runs and scenarios, and train with \catrac's distributional critic and task impersonation 
-- supporting multi-modal, high-variance targets without suppressing the game's natural dynamics. 

\noindent
We use two execution modes of the game: a \emph{headless} build, which runs the full simulation without rendering (no video output), and a \emph{graphical} build, which renders frames and supports video capture. All data reported in this paper is generated by agents trained and evaluated using the headless build. The graphical build is used to visually inspect and interpret the learned behaviors, and all supplemental videos are recorded from this rendered version. Consequently, due to small dynamics discrepancies between the two builds, these videos are slightly out of distribution relative to the headless training/evaluation setting.

\paragraph*{Actions.} 
\label{sec:hrz-actions}
We interface with a \PlayStation5/ gamepad comprising 14 binary buttons, two 2D analog sticks, and two analog triggers.
To reflect human ergonomics (thumbs on sticks, D-pad, face buttons) and avoid impractical combinations, we impose \emph{virtual hand} constraints that (i) prevent simultaneous presses with no in-game effect and (ii) mirror the trade-offs of a casual player. Concretely, at most one face button and one D-pad direction may be pressed at a time; L1 and L2 are mutually exclusive, as are R1 and R2; when the right stick is active, face buttons are disabled for that step; symmetrically, when the left stick is active, L3 may be pressed but D-pad inputs are disabled. We additionally exclude the triangle button and R3, treat the triggers as binary inputs (a press corresponds to a full pull), and expose only valid trigger–button combinations (booleans) as the support of categorical heads. Continuous stick values are outputted in $[-1,1]^2$.
Even under these constraints, the action space remains large: 315 discrete combinations plus four continuous dimensions (the two sticks). The policy acts every six frames (Horizon runs near 30\,fps in our setup), so inputs are held for a 6-frame window ($\approx$5\,Hz control).

\paragraph*{Features.}
\label{sec:hrz-features}

The policy and value network inputs are the current observation and a representation of the style-request. Observations are based on data coming from the game engine (not pixels) and include features of the current step plus short histories of actions and features from previous steps. Because the raw game state is immense and often irrelevant (e.g., mesh vertices, textures), we extract a focused subset. However, some entities (e.g., transient projectiles) lack a consistent and concise representation exposed from the game engine, so a fair amount of partial observability remains.

Features cover the player, enemies, and environment. Player features include health, stamina, inventory counts, and weapon loadout. Enemy features include position/orientation, health, active attack ID (if available), and body-part descriptors (types, per-part health, current part positions)---crucial in HFW because parts affect combat mechanics (e.g., enabling abilities and attacks, exposing weak points and elemental vulnerabilities). All network inputs are scaled to lie roughly in $[-1,1]$.
To aid generalization, we use agent-centric coordinates: each 3D position is expressed relative to the player and aligned to camera axes.
The resulting relative 3D position vector \(\langle x,y,z\rangle\) is encoded as $\{x/d, y/d, z/d, \sin(d), \sin(0.1 \, d), \sin(0.01 \cdot d)\}$, where $d=|\langle x, y, z \rangle|$, describing direction (unit vector) plus a multi-scale radial encoding of the distance. 
The style request is encoded as $f(\theta_t)$ by normalizing each parameter $\theta_{i,t}$ to $\bar{\theta}_{i,t} \coloneqq \theta_{i,t}/\max\{|a_i|,|b_i|\}$ over its domain $[a_i,b_i]$, and then concatenating $[\bar{\theta}_{1,t},\ldots,\bar{\theta}_{m,t}]$. The augmented state $\tilde{s}_t=(s_t,f(\theta_t))$ exposes these normalized style requests to policy and critic.

\paragraph*{Rewards.}
\label{sec:hrz-rewards}
 A brief text description of every reward term is provided in Table~\ref{tab:all_rewards_descriptions_hfw} in Appendix \ref{app: extended data}. 
Task rewards, which are constant across styles, define the core objective. Table~\ref{tab:static_rewards_hfw} in Appendix \ref{app: extended data} lists each task reward component with its unscaled range and weight $w$. Style rewards are modulated by style parameters (as above) and are described in Table~\ref{tab:dynamic_rewards_hfw} in Appendix \ref{app: extended data}.

\paragraph*{Style Parameters.} 
\label{sec:hrz-style-parameters}
In HFW, we have $m=18$ style weights and no $\xi_i$ parameters, supporting a total of 20 style behaviors (including the `default' and no-Skykiller styles). Each weight $w_i$ lies in a continuous range, with $a_i < 0 < b_i$. Weights are sampled episodically from a mixture:
(i) with probability $0.03$, set $w_i=0$ for all $i$ (creating the default style);
(ii) with probability $0.10$, sample each $w_i \sim \mathcal{U}([a_i,b_i])$ independently;
(iii) with probability $0.87$, draw $k \sim \mathcal{U}(\{1,\dots,m\})$, sample $w_k \sim \mathcal{U}([a_k,b_k])$, and set $w_i=0$ for all $i\neq k$ (continuous one-hot). 

\paragraph*{Scenarios.} 
\label{sec:hrz-scenarios}
We train for 1-vs-1 combat across multiple enemy types and locations. We use three mostly flat sites---Ice-lake, Shining-wastes (desert land), and the in-game combat Arena, which is walled. We select \NumMachinesInUse~reliably spawning enemies from the \HFWNumMachines~machine types in the game (Table~\ref{tab:machineDetails}, Appendix \ref{app: extended data}). Aerial/aquatic or unstable enemies are excluded. The agent starts each fight at level 50 with a full skill tree, full stamina, 14 healing berries (+105 HP each), and a loadout comprising \NumTrapsInLoadout~traps of each type (acid, purgewater, shock, blast) plus \NumWeaponsInUse~weapons spanning major classes and elemental ammo (Table~\ref{tab:weaponsDetails}, Appendix \ref{app: extended data}). Valor surge is disabled and no coils or weaves are applied. For each training task we uniformly sample a location and enemy. To ensure sufficient variation, the agent and enemy spawn with random orientations within 25\,m of a manually annotated center. Episodes terminate on enemy defeat, agent defeat, stepping outside the Dynamic Anchor Bound (DAB), or reaching 4500 steps. The DAB captures the spatial relationship between the agent and nearby hostile entities by defining a bounded region whose center is either the agent's current position or a recent past position: when the agent moves toward the nearest hostile entity, DAB recenters on the agent's new position; when the agent moves away, the center remains fixed. A style request is sampled as described earlier and held fixed for the episode.

\paragraph*{Models.}
\label{sec:hrz-models}
Policy and critic are 4-layer MLPs with widths $(4096,\,2048,\,1024,\,512)$: each block is
Linear $\rightarrow$ LayerNorm $\rightarrow$ ReLU with a padded residual connection. We alternate the padding location to avoid identity paths, i.e. we use pad-at-end for layers of size
$4096$ and $1024$, and pad-at-beginning for $2048$ and $512$.
Initialization uses Glorot–Uniform weights and zero biases; networks are trained with \catrac.

\noindent\textit{Policy.}
Actions for analog sticks use tanh-Gaussian distributions, with mean $\mu\in[-1,1]$ (after tanh) and standard deviations $\sigma$ obtained by squashing the network output to $[-2,2]$ with $2\tanh(\cdot)$ and then exponentiating. The discrete actions necessary to enforce the virtual hand constraints use Categorical distributions, learned via the Gumbel-max regret matching trick mentioned in the Training Algorithm appendix. To obtain the entropy regularization term that's necessary for \catrac, empirical estimates of Squashed Gaussian actions' entropy are summed with the exact entropy values of the Categorical distribution actions; 1000 samples are used to obtain the empirical estimates. Table~\ref{tab:hfw_policy_heads} in Appendix \ref{app: extended data} reports the detailed policy head layout.

\noindent\textit{Critic.} Categorical distributional $Q$ with $255$ bins on $(-600,1000)$, trained by cross-entropy against HL-Gauss TD(0) targets ($0.75$ std/width ratio).

\noindent\textit{Model explanations.} Feature-attribution analyses provide evidence that the learned policy is sensitive to the UVFA parameters. The Saliency\citep{Simonyan14a} (gradient-based) method 
consistently ranks UVFA style inputs among the most influential features (Table~\ref{tab:feat_rankings_saliency}, Appendix \ref{app: extended data}),
while SHAP\citep{NIPS2017_7062} primarily attributes action selection to observation features capturing player and enemy state (Table~\ref{tab:feat_rankings_shap}). 
Although attribution methods do not establish causality, the consistent separation between sensitivity to conditioning variables and explanatory contributions of state features is consistent with the agent behavior modulation through UVFA parameters.

\paragraph{Computing resources.}
The trainer runs as a single process with 10.5 vCPU, 55~GiB RAM, and one NVIDIA H100 GPU, writing checkpoints every $\sim$10 minutes. We train our Horizon runs for 1500 epochs, corresponding to approximately 3 days (average training time is 470 epochs/day). Unless otherwise stated, all reported results use 100 rollout workers, although most exploratory experiments were run with 50 rollout workers and showed only marginal differences.

\subsubsection*{Gran Turismo}
We study style control for \GTSophy/ in a variety of ways: drifting with 360-degree loops (a video is available through Appendix \ref{app: videos}), strategic control with tire and fuel saving, and modulating aggressiveness by controlling the longitudinal and lateral distance to opponent cars. In each case, the designer is given control over the style to show a wide range of capabilities. Unless noted, details of features, actions, training approach, and rewards remain the same as in \cite{wurman2022outracing}.

The drifting style was showcased on the \emph{Trial Mountain} track with a Lamborghini Countach. The tire and fuel saving style control was trained on the \emph{Autodrome Lago Maggiore} track with the Red Bull X2019 Competition car.  
Aggressiveness control was studied on the \emph{24 Heures du Mans} track, while learning to drive over 500 cars and 9 tire compounds.

\paragraph*{Style reward and parameters.}
The drifting style vector is composed of $m=4$ style weights $w_i$ (no $\xi_i$), controlling the lateral slip angle, longitudinal slip ratio for the front and rear car tires. Each $w_i$ is sampled independently from a three-way mixture: $w_i\sim0.5\delta_0+0.25\mathcal{U}([-0.4, 0.0])+0.25\mathcal{U}([0.0, 5.0])$, where $\delta_0$ is a Dirac point mass at $0$. 

For tire and fuel saving, we use $m=4$ style weights $w_i$ (no $\xi_i$), to control overtaking, defending, tire saving, and fuel saving.
We sample $w_{\text{overtake}},w_{\text{defend}}\sim\mathcal{U}([0, 2])$, $w_{\text{tire}}\sim 0.10\delta_{0} + 0.45\,\mathcal{U}([-60,0]) + 0.45\mathcal{U}([0,12000])$, and $w_{\text{fuel}}\sim0.10\delta_{0} + 0.45\mathcal{U}([-3,0]) + 0.45\mathcal{U}([0,105])$. To make these styles learnable, the observation includes tire temperature/grip and powertrain signals (RPM, torque, gear).

Finally, for aggressiveness modulation, the style vector is $(\xi_{\mathrm{lon}},\xi_{\mathrm{lat}})$ (meters), i.e. longitudinal $\xi_{\mathrm{lon}} \sim \mathcal{U}([0,2])$ and lateral $\xi_{\mathrm{lat}} \sim \mathcal{U}([0,1])$ car-body distance parameters and no weight. The sampled parameters are applied to all cars for an episode and are used to compute a dense penalty as our style reward: $P=(d_{\mathrm{lon}}-\xi_{\mathrm{lon}})_{-}+(d_{\mathrm{lat}}-\xi_{\mathrm{lat}})_{-}$ with $(x)_{-}\coloneqq\min(0,x)$, summed over nearby opponents, where $d_{\mathrm{lon}},d_{\mathrm{lat}}$ are the shortest longitudinal/lateral body distances (with cars approximated by 4-vertex polygons). Here smaller distance margins encourage more assertive passing maneuvers while larger margins produce cautious driving. Results are shown in Figure~\ref{fig:lat-control} in Appendix \ref{app: extended data}.

\paragraph*{Scenarios.}
To train the drifting style, we use similar scenarios to the ones in \citet{wurman2022outracing}. Additionally, at the start of every episode we sample style weights, and cars are launched at random positions on track with random speeds and orientations. 

To learn tire- and fuel-saving behavior, the agent must experience trajectories spanning a wide range of tire-wear levels---across 3 racing tire compounds---as well as different fuel loads. In our training scenarios, the initial tire wear conditions are sampled as $X\sim\mathcal{U}([0, 0.75])$, and for each tire we then draw the wear from $\sim N(X, 0.05)$. The initial fuel level is instead sampled from $\mathcal{U}([10.0, 100.0])$.

Training the aggressiveness style uses cars drawn from a pool of $530$ vehicles, with adaptive sampling that up-weights cars prone to spinning out or losing control. Tires are sampled uniformly from $9$ compounds (comfort/sport/race $\times$ hard/medium/soft). During evaluation we focus on the GT3-class of cars and report averages over multiple races for each of the $50$ Gr.3 cars. Episodes use rolling starts with $5$ cars of the same model, over $4$ laps, using a combination of fixed and sampled distance parameters to analyze the learned behaviors. The evaluations match the approach taken in \cite{werner2023dynamic} with agents starting in leading and pursuing positions. For each race, we log the win-rate both as a leader and pursuer, overtakes, and collision events.

\subsubsection*{DMC Humanoid}
We apply style-conditioned UVFA in the \emph{Humanoid} task from the \href{https://github.com/google-deepmind/dm_control/tree/main/dm_control/suite}{\emph{DeepMind Control Suite}} (DMC)~\citep{tunyasuvunakool2020dm_control,todorov2012mujoco}, where the agent controls a simulated humanoid on an infinite plane. 
The core task is forward walking; styles target combinations of arm poses and gait.

\paragraph*{Actions.} The \texttt{walk} task requires the humanoid to stand up and then move at a minimum speed of 1 over episodes of 1000 time steps. The 21-dimensional continuous action space $A \coloneq [-1, 1]^{21}$ specifies control inputs for each joint actuator. 

\paragraph*{Features.} 
Observations are proprioceptive and include angle and velocity of the joints, as well as velocity of the center of mass. Additionally, we include features to support the gait style: one-hot of the foot that should strike next; per-step min/max knee angles (both legs); normalized pelvis orientation; and relative foot positions $(x,y)$ recorded at the start of the current step and stride. 

\paragraph*{Rewards:} 
The total reward decomposes into a core movement term $r_{task} = r_{\mathrm{move}}$ and two style terms ($r^{\sigma_i}$):
\[
r = r_{\mathrm{move}} + \lambda_{\mathrm{pose}}\, r_{\mathrm{pose}} + \lambda_{\mathrm{gait}}\, r_{\mathrm{gait}},
\]
with $\lambda$ fixed multipliers (not style parameters $\theta$). At every step, $r_{\mathrm{move}}$ incentivizes \emph{forward} motion only, while penalizing large torques and decaying as head height drops.
Both style terms use DMC \texttt{tolerance()} with \texttt{long\_tail} decay (dense gradients, bounded in $[0,1]$ before shifting/scaling). At every step, the \emph{arm pose reward} ($r_{\mathrm{pose}}$) keeps six arm joints (\texttt{ELBOW}, \texttt{SHOULDER1}, \texttt{SHOULDER2} per arm) within pose-specific angles for $\xi_{\mathrm{pose}}$---with margin $\approx 30^\circ$ around bounds. The \emph{gait reward} ($r_{\mathrm{gait}}$) instead targets a step length $\xi_{\mathrm{gait}} = \ell$ on alternating steps using 
margin $m=0.05$.
Here we use additional shaping terms: (i) stride contribution bounds $[1/3,\,2/3]$ per foot to avoid ``vibrating'' gaits; (ii) knee range-of-movement limits (min/max knee angles within $[-145^\circ,0^\circ]$); (iii) timeout penalty $-v_m$ if no alternation for $t_{\text{step}}>t_{\text{step\_max}}$.

\paragraph*{Style Parameters.} Humanoid conditions on a style vector \(\theta=(\xi_{\mathrm{pose}},\xi_{\mathrm{gait}})\).
These style-specific parameters are independently sampled at each episode and encode the style geometry and constraints:
\(\xi_{\mathrm{pose}}=\text{onehot(pose\_id)}\), where \(\text{pose\_id}\in\{\text{arms-in-front},\text{arms-up},\text{T-pose}\}\) corresponds to specific arm joint-angle bounds with \(\sim30^\circ\) margins; 
\(\xi_{\mathrm{gait}}=\ell\), where $\ell$ is the target step length. Therefore, arm poses and gait constraints are trained jointly.

\paragraph*{Scenarios.} We modify initialization/termination to speed early learning on \texttt{walk}: the humanoid starts upright (non-arm joints at $0^\circ$, arm joints randomized).
An episode terminates when head height $<1.1$ (i.e., a fall) or at $t=1000$.  A style request is sampled as described earlier. 

\paragraph*{Models.} Policy and critic are 3$\times$512 residual MLPs with LayerNorm, trained using SAC with Adam (lr $10^{-4}$), discount $0.99$, and fixed entropy weight $\alpha=0.01$.
\textit{Policy:} $\tanh$-Gaussian over the 21-D torque actions.
\textit{Critic:} twin scalar $Q$-nets with Polyak-averaged targets.

\paragraph*{Computing resources.}
Humanoid experiments ran on a single NVIDIA H100 GPU with 10 CPU rollout workers and experience replay (buffer size $7.5\times10^{5}$ transitions).
Training was throttled until 15000 steps were collected and further constrained to 60 env-steps/s.


\clearpage
\section{Extended Data}
\label{app: extended data}

\begin{table}[ht]
\centering
\caption{Per-domain training configuration.}
\label{tab:domain_training_summary}
\small
\setlength{\tabcolsep}{4pt}
\renewcommand{\arraystretch}{1.2}
\begin{tabular}{@{} L{2.7cm} L{4.6cm} L{4.6cm} L{4.6cm} @{}}
\toprule
 & \textbf{Humanoid (DMC)} & \textbf{Gran Turismo 7} & \textbf{Horizon Forbidden West (HFW)} \\
\midrule
\textbf{Algorithm} &
SAC~\citep{haarnoja2018soft} &
QR-SAC~\citep{wurman2022outracing} &
Categorical Regularized Actor–Critic (Cat-RAC) \\
\addlinespace
\textbf{Critic / Value} &
Twin $Q$ (min over 2) &
Distributional critic (quantile regression) &
Categorical value (255 bins); HL-Gauss TD(0) targets \\
\addlinespace
\textbf{Update order} &
Simultaneous actor/critic &
Policy then critic &
Policy then critic \\
\addlinespace
\textbf{Action space} &
Continuous ($\tanh$-Gaussian) &
Continuous &
Mixed: categorical buttons + $\tanh$-Gaussian sticks/triggers \\
\addlinespace
\textbf{Regularization} &
Entropy; fixed $\alpha$ &
Entropy (SAC-style) with distributional $Q$ &
Exact categorical entropy; Gaussian entropy via $n{=}10^3$ samples \\
\addlinespace
\textbf{Task impersonation} &
No & No & Yes \\
\addlinespace
\textbf{Stratified Sampling} &
No & Yes (multi-tables\citep{wurman2022outracing}) & Yes (event tables\citep{kompella2023eventtables}) \\
\addlinespace
\textbf{Scenario training} &
Yes & Yes & Yes \\
\bottomrule
\end{tabular}
\end{table}

\clearpage

\clearpage

\begin{figure}[ht]
\centering
\includegraphics[width=\linewidth]{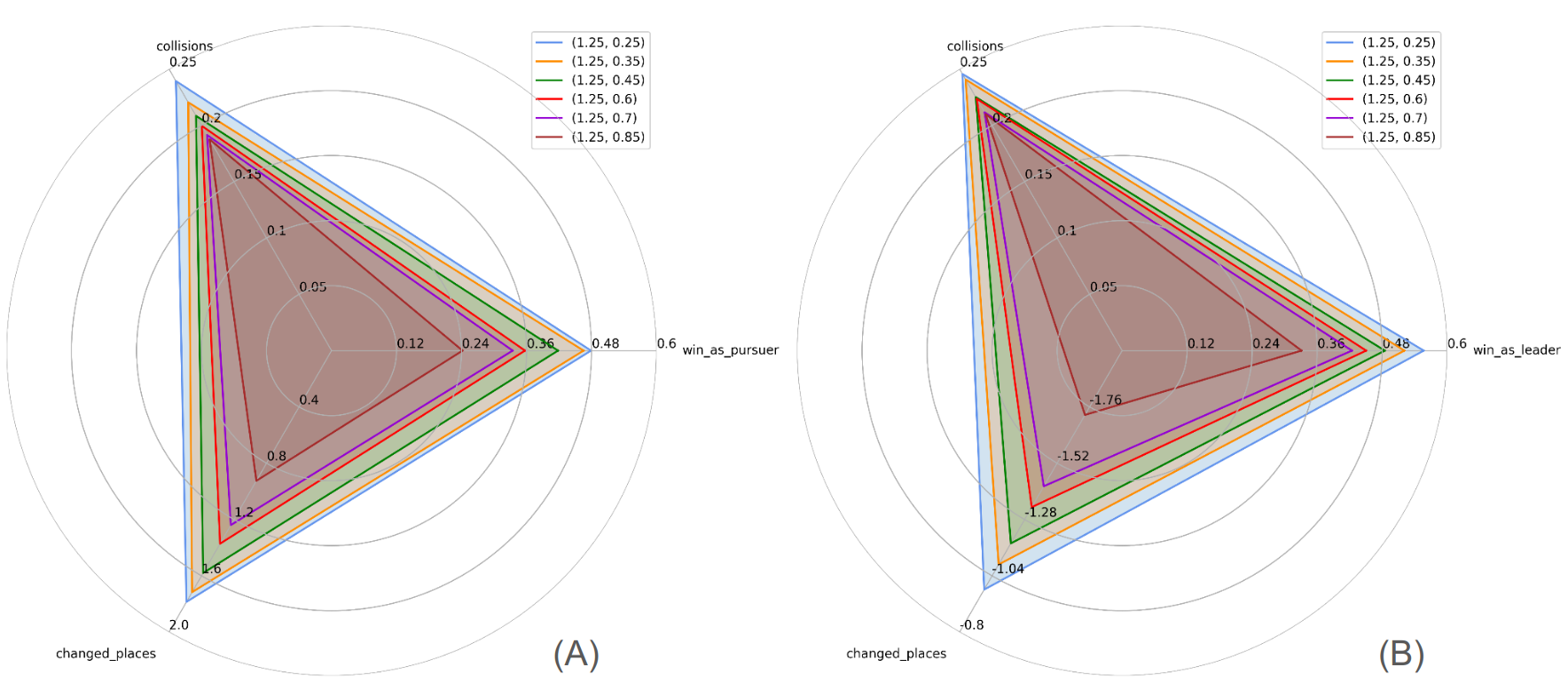}
\caption{
\textbf{Controlling the level of aggressiveness:} \GTSophy/ can be controlled to drive more or less aggressively by adjusting the longitudinal and lateral distance margins and penalizing the agent when it violates the margins. Lower margins allow the agent to drive closer to the opponent cars, facilitating pass attempts and overtakes, and resulting in more competitive driving. Conversely, higher margins induce more timid behavior, with the agent maintaining larger distances to opponents and backing away when approached by an opponent car, which makes overtaking more difficult. Radar plots (A) and (B) report metrics for 5-car races in a \textit{pursuer} and \textit{leader} setting respectively. In the pursuer scenario two agents start at the back of the group (positions 3 and 4), while in the leader scenario they start at the front (starting positions 0 and 1). In both settings, the two agents are assigned target margins, while the opponent cars occupying the remaining three positions use permutations of (0.5, 0.75) as lateral margins and share the same longitudinal margin. These plots show how the number of rank changes (i.e., positions gained or lost), the percentage of cars finishing in leading positions (win rates), and the number of collision events scale as the lateral distance margins are varied. This experiment uses launch configurations and a win rate metric inspired by \cite{werner2023dynamic}. Plot (A) shows that agents with lower lateral margin parameters gain more positions towards the finish line, increasing their win rate as a pursuer, but at the cost of colliding more with opponents. In contrast, plot (B) demonstrates that agents using lower lateral margins win more frequently in the leader scenario by reducing the number of positions lost during the race; they defend their leading positions more aggressively by blocking opponents, again at the expense of increased collisions. In both plots, increasing the lateral margins results in less competitive behavior, with agents maintaining larger inter-car distances, leading to fewer overtakes, fewer collisions, and lower win rates.
}
\label{fig:lat-control}
\end{figure}

\clearpage

\begin{table}[ht]
    \centering
    \caption{The Horizon Forbidden West machine types used in training and testing. Health indicates how much damage a machine can take. A machine strong to an elemental type takes half damage, while a machine weak to an elemental type takes twice the damage. The column ``Parts: S, W, C'' shows the number of removable shield, weapon, and canister parts that the machine has, respectively.
    Overall, there are significant challenges with partial observability, with most enemies having at least one attack that is partially or entirely unobserved.
    }
    \label{tab:machineDetails}
    \begin{tabular}{|l|c|l|l|l|} \hline
Machine & Health & Strengths & Weaknesses & Parts: S/W/C \\ \hline 
Burrower & 95 & none & fire & 0, 1, 1 \\ \hline 
Scrounger & 150 & none & frost, acid & 0, 1, 1 \\ \hline 
Scrapper & 235 & plasma & shock & 1, 1, 1 \\ \hline 
Leaplasher & 375 & shock & purgewater & 0, 2, 2 \\ \hline 
Lancehorn & 390 & frost & fire & 2, 2, 4 \\ \hline 
Longleg & 750 & acid & shock & 2, 3, 2 \\ \hline 
Widemaw & 900 & acid, frost, fire, & purgewater & 0, 3, 6 \\
& & plasma, shock & & \\ \hline 
Ravager & 1300 & frost, shock & acid, purgewater & 1, 1, 3 \\ \hline 
Snapmaw & 1450 & frost, purgewater & fire, shock & 1, 0, 15 \\ \hline 
Grimhorn & 1800 & shock, fire & purgewater, acid & 9, 6, 2 \\ \hline 
Scorcher & 2000 & fire & frost, shock & 1, 3, 3 \\ \hline 
Rollerback & 2200 & frost, shock & acid & 4, 6, 8 \\ \hline 
Specter & 2400 & shock, purgewater, fire & acid, plasma & 12, 12, 8 \\ \hline 
Frostclaw & 2800 & frost & fire, shock & 3, 2, 5 \\ \hline 
Rockbreaker & 3600 & fire & frost, shock & 2, 6, 2 \\ \hline 
Slitherfang & 5500 & shock, purgewater, acid & frost, plasma, fire & 28, 7, 24 \\ \hline 
Apex Thunderjaw & 6500 & fire, frost, shock, & acid & 11, 5, 22 \\
& & purgewater, plasma & &  \\ \hline 
Slaughterspine & 7500 & acid, plasma, shock, fire & purgewater, frost & 2, 12, 17 \\ \hline 
Specter Prime & 8500 & shock, purgewater, fire & acid, plasma & 18, 18, 8 \\ \hline 
        \end{tabular}
\end{table}
\clearpage

\begin{table}[ht]
    \centering
    \caption{Aloy's weapons configured in the experiments. We chose examples from six different weapon classes to create a variety of different weapon styles, providing the agent with a variety of elemental attacks. To change weapons, the agent must hold the L1 button on the controller and then use the right joystick to point in the desired direction on the wheel. Each sixth of the wheel corresponding to a weapon is further divided into two or three ammo subsections, depending on the type of weapon. Note that the spear, used for melee attacks, is always available and not part of the weapon wheel.}
    \label{tab:weaponsDetails}
    \begin{tabular}{|l|l|l|} \hline
Weapon (Type) & Ammo & Comments\\ \hline 
Icestorm & Frost Bolts &  \\ 
(bolt blaster) & Plasma Bolts & \\ 
  & Shock Bolts &  \\ \hline 
Carja's Bane & Light Arrow &  \\
(warrior bow) & Fire Light Arrow &  \\ \hline 
Corrosive Blastsling & Acid Bomb &  \\ 
(blastsling) & Adhesive Bomb &  \\ 
  & Purgewater Bomb &  \\ \hline 
Sun Scourge & Advanced Fire Hunter Arrow &  \\
(hunter bow) & Advanced Frost Hunter Arrow &  \\ 
  &Advanced Acid Hunter Arrow  &  \\ \hline 
Skykiller & Advanced Explosive Spike &  \\ 
(spike thrower)  & Explosive Spike &  \\ 
  & Fire Spike &  \\ \hline 
Ancestor's Return & Acid Shredder &  \\
(shredder gauntlet) & Tear Shredder &  \\ 
  & Shock Shredder &  \\ \hline 
    \end{tabular}
\end{table}

\clearpage

\begin{table}[ht]
    \centering
    \caption{The agent's aggregate record against all of the machines, in the format: \textit{wins-loses-out of bounds-timeouts}. The machines are sorted from the weakest to strongest in terms of machine health. As expected, the agent's performance drops off against the stronger enemies, though even against the toughest opponent, the Slaughterspine, the agent wins about 60\% of the time. Interestingly, the agent has a surprisingly large number of 20 minute timeouts against the three weakest enemies, which we attribute to the value for removing parts being significantly more than the health of the enemy, causing the agent to chase the machine around looking for the perfect shot.}
    \label{tab:enemyRecords}
    \begin{tabular}{|l|l|} \hline
        \textbf{Machine} & \textbf{Record} \\ \hline 
        Burrower  & 1083-0-0-57 \\ \hline 
        Scrounger  & 1129-1-0-10 \\ \hline 
        Scrapper  & 1122-0-0-18 \\ \hline 
        Leaplasher  & 1135-1-0-5 \\ \hline 
        Lancehorn  & 1140-0-0-0 \\ \hline 
        Longleg  & 1134-1-6-0 \\ \hline 
        Widemaw  & 1132-3-2-6 \\ \hline 
        Ravager  & 1134-8-0-0 \\ \hline 
        Snapmaw  & 1125-21-0-0 \\ \hline 
        Grimhorn  & 1135-4-0-3 \\ \hline 
        Scorcher  & 1113-26-0-4 \\ \hline 
        Rollerback  & 1135-5-1-0 \\ \hline 
        Specter  & 1135-7-1-0 \\ \hline 
        Frostclaw  & 1008-122-13-2 \\ \hline 
        Rockbreaker  & 1054-58-1-28 \\ \hline 
        Apex Thunderjaw  & 767-350-1-24 \\ \hline 
        Slitherfang  & 1090-46-5-0 \\ \hline 
        Slaughterspine  & 656-482-3-0 \\ \hline 
        Specter Prime  & 967-108-1-66 \\ \hline
    \end{tabular}
\end{table}

\clearpage

\clearpage

\begin{table}[ht]
\centering
\caption{
\textbf{HFW Style Score and Win Rate. }
The data is collected from fights against all 19
enemies 10 times in each of the three test locations.
}
\label{tab:baseline style score win rate}
\small
\setlength{\tabcolsep}{6pt}
\renewcommand{\arraystretch}{1.15}
\begin{tabular}{@{} l c c @{}}
\toprule
\textbf{Style} & \textbf{Style score} & \textbf{Win rate} \\
\midrule
Default Behavior & ---  & 1.00 \\
Melee & 0.90 & 0.83 \\
Traps & 0.98 & 0.92 \\
\midrule
Weapons: Ancestor's Return  & 1.00 & 0.98 \\
Weapons: Carja's Bane  & 0.98 & 0.98 \\
Weapons: Corrosive Blastsling  & 1.00 & 0.98 \\
Weapons: Icestorm Boltblaster  & 1.00 & 0.88 \\
Weapons: Skykiller  & 1.00 & 0.99 \\
Weapons: Sun Scourge  & 1.00 & 0.93 \\
\midrule
Elementals: Acid & 0.84 & 0.98 \\
Elementals: Adhesive & 0.84 & 0.94 \\
Elementals: Fire & 0.84 & 0.98 \\
Elementals: Frost & 0.85 & 0.99 \\
Elementals: Plasma & 0.48 & 0.94 \\
Elementals: Purgewater & 0.73 & 0.97 \\
Elementals: Shock & 0.41 & 0.99 \\
\midrule
Parts: Canister & 0.37 & 0.96 \\
Parts: Shield & 0.60 & 0.97 \\
Parts: Weapon & 0.44 & 0.94 \\
\bottomrule
\end{tabular}
\end{table}

\clearpage

\clearpage

\clearpage

\clearpage

\clearpage

\clearpage

\begin{table}[ht]
\centering
\caption{Descriptions of all rewards for HFW.}
\label{tab:all_rewards_descriptions_hfw}
\begin{tabular}{ | m{14em} | m{24em} | }
\hline
Reward & Description \\
\hline\hline

Damage dealt & Damage dealt to the enemy from any source \\
\hline
Show weapon wheel & Penalty when the weapon wheel is displayed \\
\hline
Show quick menu & Penalty for opening the quick menu \\
\hline
Out of bounds & Penalty for leaving the anchored combat area \\
\hline
Player death & Penalty for player death \\
\hline
Enemy off screen & Penalty for enemies being off screen \\
\hline
Right joystick smoothness & Penalty for non-smooth joystick movement (right joystick only) \\
\hline
Trap damage dealt  & Damage dealt to the enemy from traps \\
\hline
Trap proximity & Penalty for the player's proximity to active traps \\
\hline
Enemy trap lure & Reward for luring the enemy into active traps \\
\hline
Weapon damage dealt & Damage dealt to the enemy from using the specified weapon \\
\hline
Target part damage dealt & Damage dealt to an enemy's part of the specified type \\
\hline
Target part removal & Binary reward for removing an enemy's part of the specified type \\
\hline
Elemental state damage dealt & Damage dealt to the enemy from any source while the enemy is in the specified elemental state \\
\hline
Activate elemental ammo & Binary reward for selecting ammo that has the specified elemental effect \\
\hline
Elemental state buildup & Reward for the specified elemental state building up on the enemy \\
\hline
\end{tabular}
\end{table}

\clearpage

\begin{table}[ht]
\centering
\caption{Task rewards for HFW. Rewards marked by $\dagger$ do not have a well defined bound; a typical upper bound estimate is given.}
\label{tab:static_rewards_hfw}
\begin{tabular}{ | m{12em} | m{8em}| m{8em} | }
\hline
\textbf{Reward name} & \textbf{Unscaled \mbox{reward} range} & \textbf{Reward weight} \\
\hline\hline

Damage dealt & $[0, 10^3]^\dagger$ & $0.03$ \\ 
\hline
Show weapon wheel & $\{0, 1\}$ & $-0.5$ \\
\hline
Show quick menu & $\{0, 1\}$ & $-0.5$ \\
\hline
Out of bounds & $\{0, 1\}$ & $-500$ \\
\hline
Player death & $\{0, 1\}$ & $-500$ \\
\hline
Enemy off screen & $[0, 1.87]$ & $-1$ \\
\hline
Right joystick smoothness & $[0, 5.66]$ & $-0.01$ \\
\hline
\end{tabular}
\end{table}

\clearpage


\newcommand{\RewardBox}[1]{\parbox[t]{10em}{\vspace{0pt}\raggedright\strut #1\strut}}
\newcommand{\RangeBox}[1]{\parbox[t]{6em}{\vspace{0pt}\raggedright\strut #1\strut}}
\newcommand{\WeightBox}[1]{\parbox[t]{3em}{\vspace{0pt}\raggedright\strut #1\strut}}

\begin{table}[ht]
\centering
\caption{Style rewards for HFW. Rewards marked by $\dagger$ do not have a well defined bound; a typical upper bound estimate is given. $\ddagger$ Different machines have different numbers of part types; therefore the upper bound is an integer depending on the machine. Two presentation notes:
(1) The ``Style group'' column clusters variants that share structure but differ by option (e.g., the \textit{Elemental} group has separate styles for Fire, Plasma, etc.), each with its own parameterization;
(2) Each style exposes a single primary dynamic weight that controls its activation. Some styles consist of multiple style-specific reward terms; the primary weight scales all terms for that style and their (weighted) sum forms the style reward.}
\label{tab:dynamic_rewards_hfw}
\renewcommand{\arraystretch}{1.15}
\setlength{\tabcolsep}{3pt}
\small

\begin{tabular}{|>{\raggedright\arraybackslash}m{7.2em}|
                >{\raggedright\arraybackslash}m{8.2em}|
                >{\raggedright\arraybackslash}m{8.8em}|
                >{\raggedright\arraybackslash}m{10em}|
                >{\raggedright\arraybackslash}m{5.7em}|
                >{\raggedright\arraybackslash}m{4em}|}
\hline
\textbf{Style group} & \textbf{Style name} & \textbf{Style weight range} & \textbf{Reward name} & \textbf{Unscaled reward range} & \textbf{Reward weight} \\
\hline\hline

Melee combat & Melee & $[-0.05\cdot 7.5,\ 7.5]$ &
Melee damage dealt & $[0,10^3]^\dagger$ & $1$\\
\hline
\multirow{5}*{
Traps use} & \multirow{5}*{Traps} &
\multirow{5}*{$[-0.05\cdot 0.375,\ 0.375]$} &
\RewardBox{Trap damage dealt\\[0.35em]Trap player proximity\\[0.35em]Trap lure enemy\\} &
\RangeBox{$[0,10^3]^\dagger$\\[0.35em]$[-1.0,1.0]$\\[0.35em]$[-1.0,1.0]$} &
\WeightBox{$1$\\[0.35em]$-100$\\[0.35em]$500$}
\\
\hline

\multirow{6}{7.2em}{Target weapon use} &
Ancestor's Return & $[-0.05\cdot 1.5,\ 1.5]$ &
\multirow{6}{10em}{\RewardBox{Weapon damage dealt}} &
\multirow{6}{6em}{\RangeBox{$[0,10^3]^\dagger$}} &
\multirow{6}{3em}{\WeightBox{$1$}}  \\
\cline{2-3}
& Corrosive Blastsling & $[-0.05\cdot 1.8,\ 1.8]$ & & & \\
\cline{2-3}
& Carja's Bane & $[-0.05\cdot 2.25,\ 2.25]$ & & & \\
\cline{2-3}
& Sun Scourge & $[-0.05\cdot 7.5,\ 7.5]$ & & & \\
\cline{2-3}
& Icestorm Boltblaster & $[-0.05\cdot 30.0,\ 30.0]$ & & & \\
\cline{2-3}
& Skykiller & $[-0.3,\ 0.3]$ & & &\\
\hline

\multirow{3}{7.2em}{Enemy part targeting} &
Weapon & $[-0.05\cdot 1.5,\ 1.5]$ &
\multirow{3}{10em}{\RewardBox{Target part damage dealt\\[0.35em]Target part removal\\}} &
\multirow{3}{6em}{\RangeBox{$[0,10^3]^\dagger$\\[0.35em]$\{0,1,2,\dots\}^\ddagger$}} &
\multirow{3}{3em}{\WeightBox{$1$\\[0.35em]$1000$}}
\\
\cline{2-3}
& Canister & $[-0.05\cdot 1.5,\ 1.5]$ & & & \\
\cline{2-3}
& Shield & $[-0.05\cdot 1.5,\ 1.5]$ & & & \\ [0.7em] 
\hline

\multirow{7}{7.2em}{Elemental} &
Acid & $[-0.05\cdot 3.0,\ 3.0]$ &
\multirow{7}{10em}{\RewardBox{Elemental state damage dealt\\[0.35em]Activate elemental ammo\\[0.35em]Elemental state buildup}} &
\multirow{7}{6em}{\RangeBox{$[0,10^3]^\dagger$\\[1.45em]$[-2.0,2.0]$\\[0.35em]$[-0.1,1.0]$}} &
\multirow{7}{3em}{\WeightBox{$1$\\[1.45em]$20$\\[0.35em]$\{15,35\}$}}
\\
\cline{2-3}
& Purgewater & $[-0.05\cdot 3.0,\ 3.0]$ & & & \\
\cline{2-3}
& Frost & $[-0.05\cdot 3.0,\ 3.0]$ & & & \\
\cline{2-3}
& Shock & $[-0.05\cdot 3.0,\ 3.0]$ & & & \\
\cline{2-3}
& Plasma & $[-0.05\cdot 3.0,\ 3.0]$ & & & \\
\cline{2-3}
& Fire & $[-0.05\cdot 4.5,\ 4.5]$ & & & \\
\cline{2-3}
& Adhesive & $[-0.05\cdot 4.5,\ 4.5]$ & & & \\
\hline

\end{tabular}
\end{table}

\clearpage

\begin{table}[ht]
\centering
\caption{Policy head outputs used in HFW.}
\label{tab:hfw_policy_heads}
\small
\setlength{\tabcolsep}{6pt}
\renewcommand{\arraystretch}{1.15}
\begin{tabular}{@{} l l l l @{}}
\toprule
\textbf{Component} & \textbf{Head / Params} & \textbf{Support} & \textbf{Dim.} \\
\midrule
(i) Sticks $(x,y)$ & 4 Squashed Gaussians & $\mu\in[-1,1]$; $\sigma=\exp(2\tanh(\,\cdot\,))$ & 8 \\
(ii) Face buttons & Categorical & $\{\varnothing,\ \text{Cross},\ \text{Circle},\ \text{Square}\}$ & 4 \\
(iii) D-pad + left analog & Categorical & $\{\varnothing,\ \uparrow,\ \downarrow,\ \leftarrow,\ \rightarrow,\ \text{L3}\}$ & 6 \\
(iv) Left shoulder & Categorical & $\{\varnothing,\ \text{L1},\ \text{L2}\}$ & 3 \\
(v) Right shoulder & Categorical & $\{\varnothing,\ \text{R1},\ \text{R2}\}$ & 3 \\
(vi) Left-side mode & Categorical & $\{\text{Thumb on Buttons},\ \text{Thumb on Joystick}\}$ & 2 \\
(vii) Right-side mode & Categorical & $\{\text{Thumb on Buttons},\ \text{Thumb on Joystick}\}$ & 2 \\
\bottomrule
\end{tabular}
\end{table}

\clearpage

\begin{table}[ht]
\centering
\caption{Ranked feature importance across 100 gameplay trajectories using SHAP and saliency. SHAP estimates each feature’s average marginal contribution to the policy’s outputs across trajectories, indicating that game state features primarily drive action selection. Saliency quantifies the local gradient-based sensitivity of the outputs to input features, revealing strong responsiveness to style parameters.}
\label{tab:feat_rankings_two_methods}
\setlength{\tabcolsep}{4pt}
\renewcommand{\arraystretch}{1.05}
\small

\begin{subtable}[t]{0.49\linewidth}
\centering
\caption{SHAP}
\label{tab:feat_rankings_shap}
\begin{tabular}{@{}r L{4.9cm} r r@{}}
\toprule
\textbf{Rank} & \textbf{Feature name} & \textbf{Mean} & \textbf{Std} \\
\midrule
1  & Enemy health & 0.90 & 0.06 \\
2  & Weapon technique stamina & 0.89 & 0.11 \\
3  & Player pose & 0.86 & 0.11 \\
4  & Enemy elemental status & 0.86 & 0.12 \\
5  & Traps: pose + inventory & 0.86 & 0.10 \\
6  & Enemy parts health & 0.84 & 0.06 \\
7  & Enemy attack state & 0.84 & 0.08 \\
8  & Player game state & 0.84 & 0.13 \\
9  & Enemy pose & 0.82 & 0.06 \\
10 & Quick-tool: health bucket & 0.81 & 0.08 \\
11 & Player health & 0.80 & 0.13 \\
12 & Active weapon + ammo & 0.79 & 0.17 \\
13 & Anchor pose & 0.78 & 0.08 \\
14 & Enemy parts pose & 0.78 & 0.06 \\
15 & Action history & 0.78 & 0.07 \\
16 & Active weapon technique & 0.72 & 0.06 \\
17 & Style weight: Sun Scourge & 0.72 & 0.08 \\
18 & Enemy game state & 0.70 & 0.06 \\
19 & Style weight: Melee & 0.65 & 0.12 \\
20 & Style weight: Purgewater & 0.64 & 0.07 \\
21 & Style weight: Icestorm BB & 0.63 & 0.15 \\
22 & Style weight: Ancestor's Return & 0.63 & 0.04 \\
23 & Ejectors: pose + duration & 0.62 & 0.03 \\
24 & Style weight: Adhesive & 0.62 & 0.12 \\
25 & Style weight: Corrosive BSling & 0.62 & 0.07 \\
26 & Style weight: Shock & 0.62 & 0.08 \\
27 & Weapon wheel open & 0.62 & 0.04 \\
28 & Style weight: Carja's Bane & 0.61 & 0.12 \\
29 & Style weight: Plasma & 0.61 & 0.03 \\
30 & Style weight: Part - Shield & 0.60 & 0.02 \\
\bottomrule
\end{tabular}
\end{subtable}
\hfill
\begin{subtable}[t]{0.49\linewidth}
\centering
\caption{Saliency}
\label{tab:feat_rankings_saliency}
\begin{tabular}{@{}r L{4.9cm} r r@{}}
\toprule
\textbf{Rank} & \textbf{Feature name} & \textbf{Mean} & \textbf{Std} \\
\midrule
1  & Style weight: Part - Shield & 7.70 & 16.70 \\
2  & Style weight: Melee & 7.43 & 6.51 \\
3  & Style weight: Icestorm BB & 7.12 & 6.25 \\
4  & Style weight: Acid & 7.07 & 11.90 \\
5  & Style weight: Part - Weapon & 6.58 & 14.76 \\
6  & Style weight: Sun Scourge & 5.54 & 5.83 \\
7  & Style weight: Part - Canister & 5.16 & 11.45 \\
8  & Style weight: Plasma & 5.10 & 8.71 \\
9  & Style weight: Frost & 5.07 & 5.99 \\
10 & Enemy pose & 4.71 & 2.22 \\
11 & Style weight: Fire & 4.67 & 5.37 \\
12 & Style weight: Traps & 4.37 & 5.15 \\
13 & Style weight: Adhesive & 4.23 & 5.20 \\
14 & Style weight: Shock & 4.11 & 4.86 \\
15 & Enemy game state & 4.08 & 1.57 \\
16 & Style weight: Purgewater & 4.04 & 5.02 \\
17 & Active weapon + ammo & 3.87 & 1.58 \\
18 & Style weight: Carja's Bane & 3.51 & 4.86 \\
19 & Player game state & 3.44 & 1.76 \\
20 & Style weight: Ancestor's Return & 3.41 & 3.38 \\
21 & Traps: pose + inventory & 3.11 & 0.85 \\
22 & Enemy attack state & 2.92 & 0.90 \\
23 & Style weight: Corrosive BSling & 2.88 & 3.44 \\
24 & Style weight: Skykiller & 2.85 & 6.36 \\
25 & Enemy parts pose & 2.26 & 1.04 \\
26 & Player health & 2.17 & 1.02 \\
27 & Enemy health & 1.88 & 1.49 \\
28 & Weapon wheel open & 1.80 & 0.63 \\
29 & Ejectors: pose + duration & 1.60 & 0.67 \\
30 & Active weapon technique & 1.57 & 0.29 \\
\bottomrule
\end{tabular}
\end{subtable}
\end{table}
\clearpage

\section{Videos}
\label{app: videos}

In a Google Drive folder (\href{https://drive.google.com/drive/folders/19F5uKziT_zkDurze5sy5iMFD4BHfD3lV}{link}), the following videos can be found:
\begin{itemize}
    \item Overview video of the project
    \item Video of \GTSophy/ drifting
    \item Video of humanoid walking and changing gaits and arm poses
    \item Videos of default style versus every enemy
    \item Videos of every style versus selected enemies
    \item OOD videos
\end{itemize}

\end{document}